\documentclass[lettersize,journal]{IEEEtran}
\usepackage{amsmath,amsfonts,amssymb}
\usepackage{booktabs}
\usepackage{multirow}
\usepackage{array}
\usepackage{tabularx}
\usepackage{makecell}
\usepackage{siunitx}
\usepackage{graphicx}
\usepackage{subcaption}   
\usepackage{wrapfig}
\usepackage{adjustbox}
\usepackage{tikz}
\usepackage{pgfplots}
\usepackage{algorithm}
\usepackage{algpseudocode} 
\usepackage{enumitem}
\usepackage{setspace}
\usepackage{verbatim}
\usepackage{textcomp}
\usepackage{url}
\usepackage{xcolor}
\usepackage[table]{xcolor}
\usepackage{stfloats}
\usepackage{listings}
\usepackage[most]{tcolorbox}
\usepackage{cite}
\def\BibTeX{{\rm B\kern-.05em{\sc i\kern-.025em b}\kern-.08em
    T\kern-.1667em\lower.7ex\hbox{E}\kern-.125emX}}

\newcommand{\tool}{\textsc{Latte}}

\begin{document}

\title{Latent Anchor-Driven Test Generation for Deep Neural Networks}

\author{
  Bin Duan$^{1}$,
  Matthew B. Dwyer$^{2}$,
  Guowei Yang$^{1*}$ \\
  $^{1}$School of Electrical Engineering and Computer Science, The University of Queensland, Australia  \\
  $^{2}$Department of Computer Science, University of Virginia, United States\\
  \{b.duan, guowei.yang\}@uq.edu.au, {matthewbdwyer@virginia.edu}
}

\maketitle

\begin{abstract}

Deep Neural Networks (DNNs) are increasingly being deployed in security-critical and safety-sensitive applications, which makes rigorous testing essential to identify and mitigate model weaknesses. Existing DNN testing approaches explore either the input space or a learned latent space. While latent-space generation can better maintain plausibility than direct input-space mutation, current methods still face a trade-off among exploration controllability, failure diversity, and seed-relative semantic drift.
To overcome these limitations, we propose \tool, a black-box testing framework that generates semantically proximate, diverse, and fault-revealing test cases by leveraging the latent space. Specifically, \tool\ encodes each input seed with a pre-trained VQ-VAE and performs a seed-centered, one-step latent mutation along directions defined by anchors sampled from alternative classes, followed by quantization and decoding back to the input space. This explores local neighborhoods around each seed within the learned latent manifold, resulting in a larger number and broader diversity of oracle-triggering prediction discrepancies under the same budget. We evaluated \tool\ on 5 datasets and 10 DNN models in single-model and multi-model testing scenarios. Across the evaluated datasets and models, \tool\ improves fault exposure and behavioral diversity under matched testing budgets. Under the single-model setting, it also maintains low seed-relative semantic drift with respect to the source seeds.

\end{abstract}


\section{Introduction}

\IEEEPARstart{D}{eep} neural networks~\cite{pinaya2020convolutional} (DNNs) have been widely applied in security-sensitive fields such as medical imaging~\cite{zhang2020review}, facial recognition~\cite{ben2021face}, and malware classification~\cite{ling2019deepsec}. Like traditional software, DNNs require extensive testing to ensure safe and reliable deployment~\cite{shahin2017continuous}. 
However, the mechanism of DNNs is significantly different from that of traditional software. 
DNN testing differs in that, for a well-trained DNN, failures are not caused by explicit logic mistakes, but rather result from the data-driven and probabilistic nature of the system: inputs in regions where predictions are less confident can be more sensitive to small, semantics-preserving mutations, and can therefore exhibit prediction changes more frequently~\cite{thota2020survey}. DNNs assign probabilities to all possible classes and select the class with the highest probability as a prediction~\cite{nguyen2015deep}.
Inspired by traditional software testing, DNN testing seeks to systematically generate test cases that can trigger diverse erroneous behaviors. Notably, unlike traditional software testing, DNNs do not crash on invalid inputs as long as the input format is correct (like image size); rather, they will still process and classify any such input. This means that, for DNN testing to be meaningful, generated test cases should remain semantically close to the original inputs; otherwise, the revealed failures may stem from semantic drift rather than genuinely exposing the model’s weaknesses.
Therefore, effective DNN test generation should produce follow-up inputs that remain close to their original seeds while still uncovering diverse failure modes under a limited testing budget.
It is also worth noting that DNN testing differs from adversarial attacks in both objectives and evaluation criteria. Adversarial attacks usually optimize attack success under explicit perturbation constraints, whereas DNN testing focuses on revealing fault-triggering behaviors under oracle-based criteria such as decision instability, model disagreement, and failure diversity. This distinction has been recognized in prior DNN testing methods~\cite{lee2020effective, wang2022bet, dola2024cit4dnn, weissl2025mimicry}. 

Various DNN testing methods~\cite{guo2018dlfuzz, odena2019tensorfuzz, lee2020effective, wang2022bet, pei2017deepxplore, dola2024cit4dnn, xie2019diffchaser, hu2023atom} have been proposed to identify model defects, and they can generally be categorized into white-box and black-box testing. In white-box testing, existing methods~\cite{lee2020effective, pei2017deepxplore, guo2018dlfuzz, xie2019deephunter} leverage neuron selection strategies to achieve high neuron coverage and identify test cases that may lead to errors. 
They typically require internal knowledge of the target model, including its structure and parameters. Yet, the internal model information may not always be accessible in practice, especially in privacy-sensitive testing scenarios~\cite{keskin2021cyber, agahari2022not}.
In contrast, black-box testing methods~\cite{odena2019tensorfuzz, kang2020sinvad, wang2022bet, hu2023atom, dola2024cit4dnn, weissl2025mimicry} do not require access to the internal details of the model. 
However, many of these methods employ fuzzing~\cite{odena2019tensorfuzz, guo2018dlfuzz, wang2022bet} to generate a large number of non-fault-revealing or semantically drifted candidates, which reduces testing efficiency~\cite{dola2021distribution}. 
Moreover, even though a mutation on the input seed is effective in generating fault-revealing test cases, many earlier methods directly mutate the input space, which can introduce noticeable semantic drift~\cite{odena2019tensorfuzz, guo2018dlfuzz, pei2017deepxplore, lee2020effective, xie2019diffchaser, wang2022bet, hu2023atom} and thus may generate test cases whose semantics significantly deviates from the original input~\cite{dola2021distribution}. 
With the help of generative models~\cite{higgins2017beta,kingma2013auto}, recent methods~\cite{kang2020sinvad, dola2024cit4dnn, weissl2025mimicry} generate test cases in learned latent spaces, which often yield follow-up inputs that remain closer to the original seeds than direct input-space mutation. However, effective latent-space transformation entails a fundamental trade-off: insufficient constraints may yield test cases that are either overly similar, limiting diversity and fault-revealing potential, or excessively distant from the data manifold, compromising semantic validity.
Therefore, based on the limitations of existing methods, we contend that for black-box DNN testing methods to be usable in practice, they should generate test cases that: (1) remain close to the original input in semantics; (2) exhibit diversity; and (3) reveal a large number of faults even for highly accurate models. 
Existing latent-space testing methods differ substantially in how they explore the space around each seed. SINVAD~\cite{kang2020sinvad} relies on unconstrained or boundary-oriented latent updates, Mimicry~\cite{weissl2025mimicry} performs iterative optimization-driven generation, and CIT4DNN~\cite{dola2024cit4dnn} emphasizes combinatorial coverage over latent factors under a differential-testing oracle. While these methods demonstrate the promise of latent-space testing, they do not explicitly frame test generation as controllable, seed-centered, multi-direction exploration around each source input. This leaves open the question of how to efficiently probe multiple semantically meaningful directions around a seed while balancing fault exposure, diversity, and seed-relative semantic proximity.

To address this gap, we present \tool, a seed-centric, anchor-guided testing strategy for generating diverse, fault-revealing follow-up inputs in a learned latent space. Specifically, \tool\ first uses a pre-trained VQ-VAE~\cite{van2017neural} to encode each original seed into a compact latent representation. It then samples anchors from alternative classes and explores multiple seed-centered directions by applying mutation at a controllable exploration degree. The mutated latent points are finally quantized and decoded back into the input space as candidate test inputs. The intuition behind this anchor-driven design is that directions pointing away from a seed toward alternative semantic regions may traverse neighborhoods where model decisions become less stable, while the learned latent manifold still helps keep generated inputs close to the original seed~\cite{fawzi2017robustness}. Rather than directly optimizing predictive uncertainty, \tool\ uses anchor-guided latent exploration to probe local neighborhoods around each seed.

We conduct comprehensive experiments on five datasets (MNIST, CIFAR10, ImageNet, FashionMNIST, and SVHN) and ten representative DNN models, covering both simple and large-scale architectures. We evaluate \tool\ under two widely adopted oracle settings, single-model testing and multi-model testing, and compare against existing latent-space methods, including Mimicry~\cite{weissl2025mimicry}, CIT4DNN~\cite{dola2024cit4dnn}, and SINVAD~\cite{kang2020sinvad}. Our evaluation uses failure count, seed coverage, testing efficiency, failure diversity, and semantic drift, where semantic drift is analyzed under the single-model setting. Across the evaluated settings, the results indicate that \tool\ improves fault exposure and diversity relative to the compared baselines while maintaining low seed-relative semantic drift.
Furthermore, semantic-drift analysis under the single-model setting indicates that \tool\ generates fault-revealing follow-up inputs while remaining close to their original seeds in representation space.
In summary, our main contributions are as follows:

\noindent \textbf{Testing Strategy.}
We propose \tool, a seed-centric, anchor-guided latent-space testing strategy for black-box image classifiers. \tool\ generates one follow-up input per seed-anchor pair via controllable latent displacement and supports both single-model and multi-model testing oracles.

\noindent \textbf{Empirical Characterization.}
We characterize the behavior of anchor-guided latent exploration in terms of fault exposure, behavioral diversity, testing efficiency, and semantic drift.

\noindent \textbf{Evaluation.}
We evaluate \tool\ against representative latent-space testing baselines under both single-model and multi-model settings, and further compare semantic drift against baseline-generated tests under matched budgets.

\noindent \textbf{Tool.} 
We provide an open-source implementation of \tool, supporting reproducible evaluation~\cite{Our}.


\section{Background}
\label{sec:pretrain}
\subsection{Generative Models}
Prior work has shown that learned latent representations provide a compact space for capturing salient semantic variation in complex data~\cite{bengio2013representation,gat2022latent,shen2020interpreting}. Such representations provide a structured latent domain in which follow-up test inputs can be generated more controllably than through direct input-space mutation~\cite{burgess2018understanding}. This capability to manipulate the latent space uniquely positions generative models to effectively utilize these latent representations to generate test cases that are distributional proximate to, but distinct from, the training data.

Variational Autoencoders (VAE)~\cite{kingma2013auto} are classic generative models comprising an Encoder and Decoder. The Encoder maps input samples $X$ into continuous latent vectors $Z$, which the Decoder reconstructs into $X_{\text{recon}}$. Building on this foundation, the Vector Quantized Variational Autoencoder (VQ-VAE)~\cite{van2017neural} replaces continuous latent variables with a Vector Quantizer~(VQ), mapping $Z$ to discrete codes. This design enhances representational capacity and yields strong performance on complex data such as images, audio, and video~\cite{yan2021videogpt}. 

In this work, VQ-VAE serves as a representation-and-decoding mechanism that enables structured test generation in a learned latent manifold. Our focus is not the generator itself, but how anchor-guided exploration in this latent space supports black-box testing.

\begin{figure}[t!]
  \centering
  \includegraphics[width=1\linewidth]{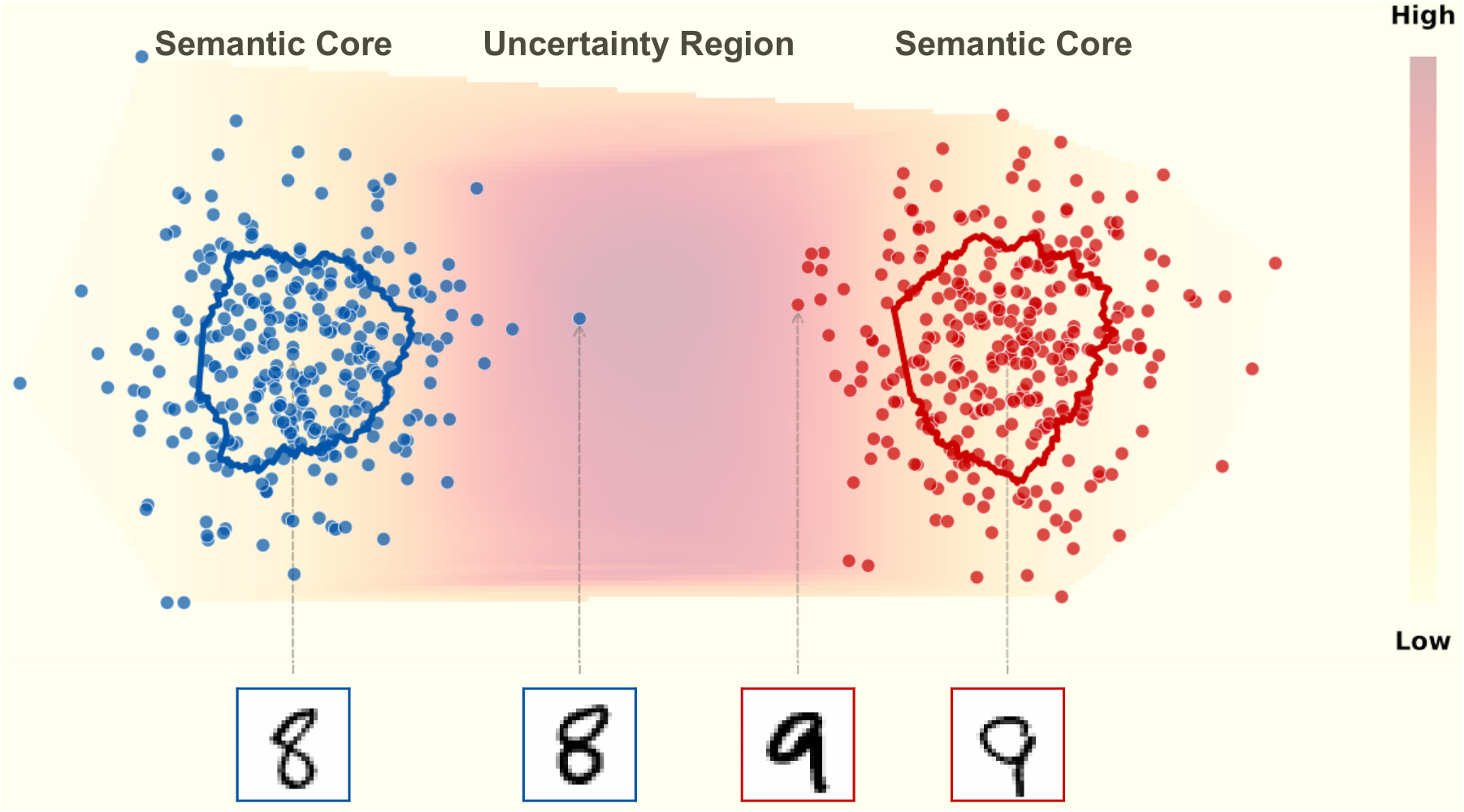}
  \vspace{-2mm}
  \caption{Latent Semantic Manifolds and Uncertainty Regions}
  \label{fig:latent_tsne}
  \vspace{-3mm}
\end{figure}

\subsection{Visualization and Rationale for Latent Space Exploration}
Figure~\ref{fig:latent_tsne} illustrates the semantic structure of a DNN model~(LeNet-4) in the latent space, visualized via t-SNE projection from the high-dimensional representation space to two dimensions. Images from different classes form distinct semantic manifolds, where blue points correspond to samples of digit~8 and red points correspond to samples of digit~9. Each class forms a compact manifold with a dense central region, referred to as the semantic core. The background heatmap visualizes the model’s predictive uncertainty at each latent location, quantified by the information entropy of the DNN’s Softmax output:
\[
H(p) = -\sum_{c \in C} p_c \log(p_c),
\]
where \(C\) denotes the set of classes and \(p_c\) is the predicted probability of class \(c\). Within the semantic core of each manifold, the model exhibits high confidence, with one class dominating the prediction (e.g., inside the blue core region, \(p_8 \approx 1\) and \(p_9 \approx 0\)), resulting in near-zero entropy. In contrast, the intermediate region between the blue and red manifolds shows substantially higher entropy, forming an uncertainty region where the classifier assigns comparable probabilities to digit~8 and digit~9 (\(p_8 \approx p_9 \approx 0.5\)), making the model more prone to misclassification. Importantly, the decoded examples suggest that many points in this intermediate region can remain visually plausible, despite exhibiting less stable predictions. This visualization highlights the contrast between semantically stable, high-confidence core regions and semantically consistent regions where model predictions become less stable. While these uncertainty patterns are illustrated for motivation purposes, they suggest that exploring regions between semantic manifolds may increase the likelihood of exposing decision instabilities, even when input semantics remain preserved.

\section{Approach}

\begin{figure*}[t!]
  \centering
  \includegraphics[width=1\linewidth]{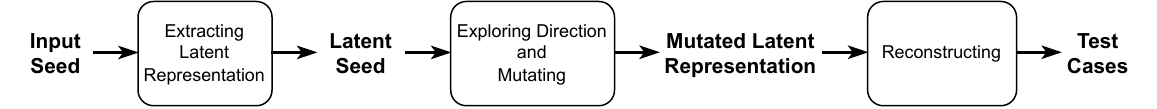}
  \vspace{-4mm}
  \caption{Overview of \tool}
  \label{fig:Overview}
  \vspace{-4mm}
\end{figure*}

The key idea of \tool\ is to perform seed-centered latent exploration using multiple anchor-defined directions.
Figure~\ref{fig:Overview} provides an overview of \tool, which consists of three main stages.

\noindent \textbf{Extracting Latent Representation:}
Given an input seed from the original dataset, \tool\ first employs a pre-trained encoder to map the input into a compact latent representation. 

\noindent \textbf{Exploring Directions and Mutating:}
Within the latent space, \tool\ explores directions from the latent seed toward different anchors; for each anchor, it applies a single mutation with a controlled exploration degree to generate a mutated latent representation that remains in the seed’s local semantic neighborhood.

\noindent  \textbf{Reconstructing:}  
Each mutated latent representation is then passed through quantizer and decoder, which reconstructs it back into the input space. This reconstruction transforms the latent representations into input test cases. 

The generated test cases are finally fed into the target DNN model. To comprehensively test DNN models, we evaluate the test cases using two oracle mechanisms: single-model testing and multi-model testing. 

\subsection{Extracting Latent Representation}
\label{sec:3.1}

The first step in \tool\ is to map input data into a structured latent space, which provides a compact, disentangled representation for subsequent exploration. Given an input seed $X_{seed}$, we employ a pre-trained VQ-VAE encoder to obtain its latent seed representation:

\begin{equation}
Z_{seed} = \text{\(Encoder\)}(X_{seed})
\end{equation}

This mapping transforms high-dimensional, pixel-level information into a lower-dimensional, semantically meaningful latent vector $Z_{seed}$ for the following mutation. Compared to mutating directly in the input space, manipulating test cases in the latent space has two major advantages: 
(1) it leverages the autoencoder’s learned latent manifold, so that mutations remain within the region of plausible samples and the decoded test cases are proximate to the original semantics; and (2) it disentangles class-relevant factors of variation, so that the following mutations align the seed’s class-relevant semantics, enabling controlled, targeted exploration along specific directions.

\subsection{Exploring Directions in the Latent Space}
\label{sec:3.2}

Once the latent representation of the input seed has been obtained, the next step is to  explore different directions in the latent space. To achieve this, we introduce the notion of anchors: for each class distinct from that of the seed, we designate many latent representations as anchors in the latent space. These anchors serve as directions for mutation, guiding the exploration around the seed toward anchors.

To this end, for a given $Z_{seed}$ belonging to class $C_{seed}$, we collect a set of anchor latent representation $\{Z_{anchor}^{(k)}\}$, each derived from an input $X_{anchor}^{(k)}$ sampled from a distinct class $C_k \neq C_{seed}$:
\begin{equation}
Z_{anchor}^{(k)} = \text{\(Encoder\)}(X_{anchor}^{(k)})
\end{equation}
These $Z_{anchor}^{(k)}$ serve as anchors distributed across the latent space, representing diverse exemplars from different classes. For a seed with class label $C_{seed}$, anchors are sampled only from classes $C_k \neq C_{seed}$. Unless otherwise stated, we use class-balanced anchor sampling: we first uniformly sample a non-seed class and then uniformly sample one anchor instance from that class. This process is repeated until the required number of seed-anchor pairs is obtained. 

The benefits of exploring along these directions are twofold:
First, exploring from a seed toward anchors from other classes encourages exploration toward regions where predictions can become less stable, which increases the chance of exposing decision instabilities under small, semantics-preserving changes.
Second, selecting multiple anchors for each class enables exploration of diverse directions as well as variations within the original seed class manifold, encouraging that the generated test cases encompass a rich variety of representative scenarios.

By operating in the latent space, \tool\ enables anchor-driven, targeted exploration that forms the basis for mutation and diverse test generation in the subsequent step.

\subsection{Mutating in the Latent Space}
\label{sec:3.3}

Building on the anchor directions established in Section~\ref{sec:3.2}, \tool\ performs mutation in the latent space to generate a diverse set of candidate test cases. Although the operator depends on both a seed and an anchor, it does not recombine latent substructures from two parents as in genetic crossover. The anchor only provides a direction, while the mutated point remains centered on the original seed.

\noindent \textbf{Mutation Strategy.}
Given a latent code $Z_{seed}$ for an input seed, \tool\ identifies a set of anchor points $\{Z_{anchor}^{(k)}\}$.
For each anchor, we define a mutation direction in the latent space and discretize this direction into a finite sequence of ordered steps.
The exploration degree $E$ indexes one of these steps via the mutation operator $\mathcal{M}(Z_{seed}, Z_{anchor}^{(k)}, E)$:
\begin{equation}
Z_{mut}^{(E)} = \mathcal{M}(Z_{seed}, Z_{anchor}^{(k)}, E)
\label{eq:latent_mut_discrete}
\end{equation}
Here, $\mathcal{M}(\cdot)$ performs a direction-guided latent displacement and returns the latent representation associated with the indexed step.
By sweeping $E$, \tool\ generates anchor-directed mutations with increasing exploration depth while remaining within the seed’s local semantic neighborhood.
For each seed, multiple anchors from other classes are sampled to explore diverse directions in the latent space, enabling both effective failure discovery and diverse test generation while keeping the generated inputs close to the original seed.

\begin{figure}[t!]
  \centering
  \includegraphics[width=1\linewidth]{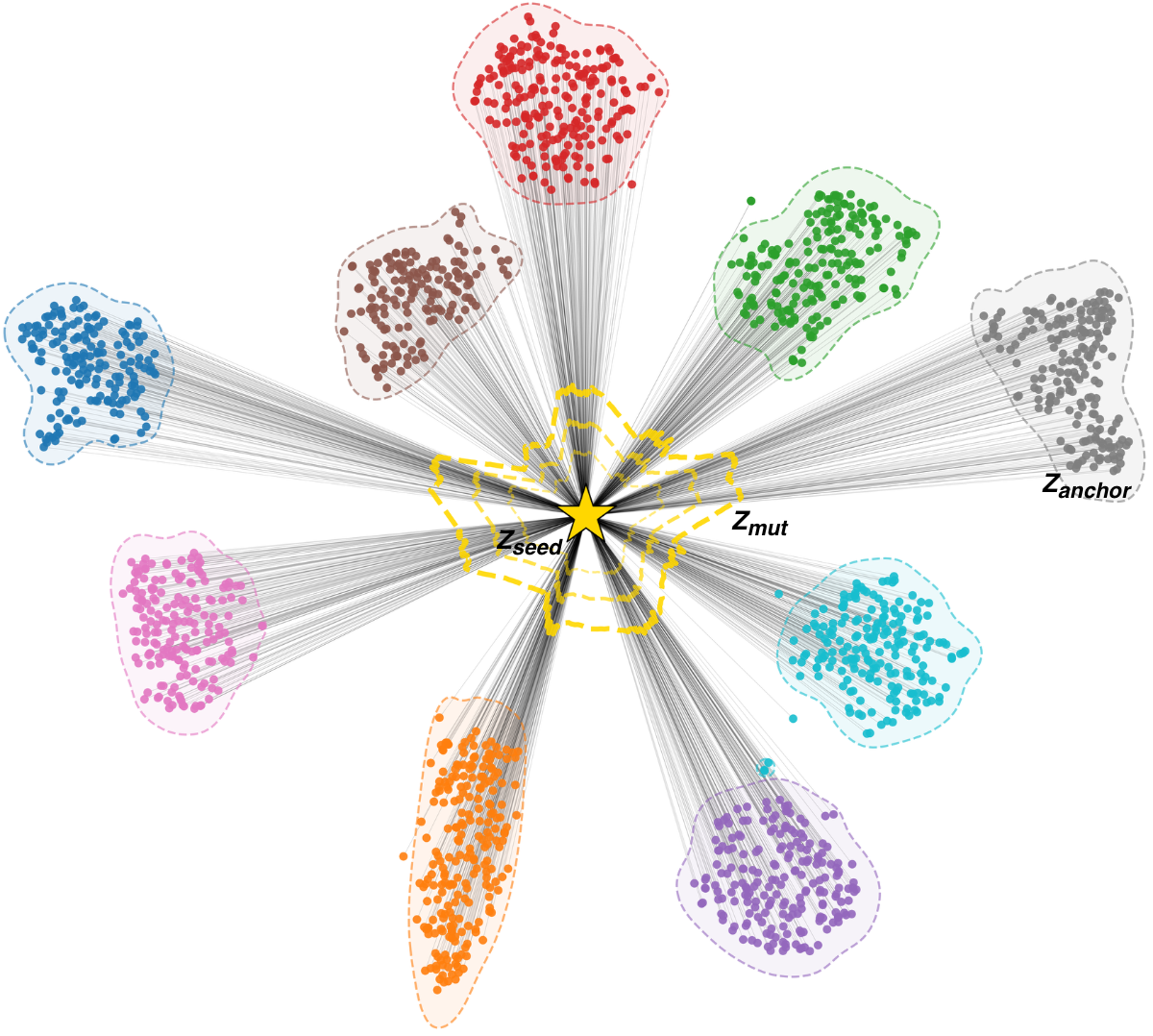}
  \vspace{-3mm}
  \caption{Mutating in the Latent Space}
  \label{fig:distri}
  \vspace{-5mm}
\end{figure}

Figure~\ref{fig:distri} illustrates the latent-space mutation mechanism of \tool\ guided by semantic manifold and model uncertainty. Given a latent seed representation \(Z_{{seed}}\) (center yellow star), \tool\ generates multiple anchors \(Z_{{anchor}}\) (other colored dots), where anchors with the same color are drawn from the same semantic region in the latent space. The vector from \(Z_{{seed}}\) to a given \(Z_{{anchor}}\) defines a mutation direction, visualized by the thin black rays radiating from the seed. The exploration degree \(E\) determines how far a mutation moves from \(Z_{{seed}}\) along the mutation direction toward \(Z_{{anchor}}\), thereby generating mutated latent representations \(Z_{{mut}}\). The yellow dashed contours around \(Z_{{seed}}\) indicate different exploration degrees. By sampling along multiple anchor directions with controlled exploration degree, \tool\ generates a diverse set of mutated latent representations that move away from the semantic core along anchor directions, thereby exploring neighborhoods where predictions may become less stable.

\subsection{Reconstructing Test Cases}
\label{sec:3.4}

After mutating seeds in the latent space, the resulting mutated latent representations should be mapped back to the input domain to yield test cases for DNN evaluation. \tool\ achieves this through a two-stage reconstruction process grounded in vector quantization and decoding with a pre-trained VQ-VAE.

\noindent \textbf{Vector Quantization.}
Given a mutated latent representation $Z_{mut}$, produced as described in Section~\ref{sec:3.3}, we first apply the VQ-VAE’s vector quantizer. This operation discretizes the continuous latent vector by mapping it to the nearest codeword in the learned codebook, enforcing that all reconstructions remain on or near the data manifold:
\begin{equation}
\hat{Z}_{mut} = \text{\(VectorQuantize\)}(Z_{mut})
\end{equation}
where $\hat{Z}_{mut}$ is the quantized latent representation, and ${VectorQuantize}(\cdot)$ denotes the quantization function, which discourages implausible test cases and ensures that generated test cases remain close to the original data manifold.

\noindent \textbf{Decoding.}
The quantized latent representation $\hat{Z}_\mathrm{mut}$ is then passed to the VQ-VAE's decoder, which reconstructs it in the input space:
\begin{equation}
\hat{X}_{test} = \text{\(Decoder\)}(\hat{Z}_{mut})
\end{equation}

This quantization-and-decoding pipeline ensures that all mutated latent representations are projected back onto the learned data manifold, effectively constraining them to remain plausible.

\subsection{Oracle.}

The final reconstructed test cases are evaluated by feeding them into the target DNN models under two widely adopted oracle settings.

\subsubsection{Single-Model Testing}
In \textit{single-model testing}, a test input is considered fault-revealing if the mutated input leads to a different prediction compared to the model's output on the original input.
This oracle, widely adopted in prior DNN testing studies~\cite{lee2020effective,pei2017deepxplore,kang2020sinvad,weissl2025mimicry}, is designed to expose {decision instabilities} of a single model under semantically proximate inputs.
Such prediction inconsistencies are not interpreted as definitive semantic errors. From a testing perspective, they constitute violations of decision stability under seed-relative, semantically proximate transformations, thereby indicating excessive sensitivity in the model’s decision behavior.

\subsubsection{Multi-Model Testing}
In \textit{multi-model testing}, a test input is considered fault-revealing if the two models in the given model pair output different predictions on the same input. This strategy, used by approaches such as~\cite{dola2024cit4dnn,wang2022bet,xie2019diffchaser}, is particularly effective for identifying corner cases and uncovering behavioral inconsistencies, especially in scenarios where ground-truth labels may be ambiguous or unavailable. 

By applying these two types of oracles, \tool\ exposes decision instabilities under single-model testing and model disagreements under multi-model testing. This dual-oracle evaluation provides a comprehensive assessment of DNN's reliability.

\subsection{Workflow of \tool}
\label{sec:3.5}

\begin{algorithm}[t]
\setstretch{1.1}
\footnotesize
\caption{Test Case Generation with \tool}
\label{alg:generate_tool}
\begin{algorithmic}[1]
\State \textbf{Input:} Dataset $\mathcal{D}$ with $K$ classes; exploration degree $E$
\State \textbf{Output:} Test case set $\mathcal{D}_{test}$
\For{each class $k$}
    \State Select a set of seeds $\mathcal{X}_{seed}^k$ from $\mathcal{D}_k$
    \State Encode $\mathcal{X}_{seed}^k$ into latent representations $\mathcal{Z}_{seed}^k$
\EndFor
\For{each $Z_{seed}^{(k,i)}$ in $\mathcal{Z}_{seed}^k$}
    \For{each target class $k' \neq k$}
        \State Select a set of anchors $\mathcal{X}_{anchor}^{k'}$ from $\mathcal{D}_{k'}$
        \State Encode $\mathcal{X}_{anchor}^{k'}$ into $\mathcal{Z}_{anchor}^{k'}$
        \For{each $Z_{anchor}^{(k',j)}$ in $\mathcal{Z}_{anchor}^{k'}$}
            \State $Z_{mut} = \mathcal{M}(Z_{seed}^{(k,i)}, Z_{anchor}^{(k',j)}, E)$
            \State $\hat{Z}_{mut} = \text{\(VectorQuantize\)}(Z_{mut})$
            \State $\hat{X}_{test} = \text{\(Decoder\)}(\hat{Z}_{mut})$
            \State $\mathcal{D}_{test} = \mathcal{D}_{test} \cup \{\hat{X}_{test}\}$
        \EndFor
    \EndFor
\EndFor
\State \textbf{Return} $\mathcal{D}_{test}$
\end{algorithmic}
\end{algorithm}
\vspace{-1mm}

Algorithm~\ref{alg:generate_tool} presents the test cases generation workflow of \tool. Given an exploration degree $E$, the process begins by partitioning the dataset $\mathcal{D}$ into $K$ class-specific subsets and encoding seeds into latent representations (lines 1–6). 

For each seed, \tool\ explores the latent space by selecting anchors from all alternative classes (lines 7–9). Each anchor defines a direction for mutation. The mutation operator $\mathcal{M}$ uses the controlled exploration degree $E$ to generate a mutated latent representation for each direction (line 12). The mutated latent representation is then quantized and decoded into the input space to obtain a candidate test case (lines 13–15). 

This workflow enables a comprehensive and controllable exploration of the latent space, enabling broad exploration of model behaviors while keeping the generated test cases close to their corresponding original seeds. By adjusting $E$, \tool\ flexibly balances between aggressive exploration and preservation of the original semantics.

\section{{Evaluation}}
\subsection{Research Questions}
We evaluate \tool\ by investigating the following research questions:
\begin{itemize}
  \item[\textbf{RQ1:}] How does \tool\ compare to existing methods in single-model testing?
  \item[\textbf{RQ2:}] How does \tool\ compare to existing methods in multi-model testing?
  \item[\textbf{RQ3:}] Does each of the main components of \tool\ contribute to its effectiveness?
\end{itemize}

For RQ1, we evaluate how \tool\ compares to existing methods when tested under the single-model testing, comparing it with state-of-the-art baselines.
For RQ2, we examine the performance of \tool\ under the multi-model testing, comparing it with state-of-the-art baselines.
For RQ3, we perform ablation experiments to investigate the effect of different exploration degrees in \tool. By varying the degree of exploration when generating mutations in the latent space, we assess how this factor influences effectiveness. Furthermore, we employ different generative models to examine how the choice of generative model impacts the results.

\begin{table*}[t!]
\centering
\small
\setlength{\tabcolsep}{4.3pt}
\renewcommand{\arraystretch}{0.9}
\caption{Target Datasets and Models}
\vspace{-2mm}
\label{tab:models}
\begin{tabular}{lcc|cc|cc|cc|cc}
\toprule
\textbf{Dataset} 
& \multicolumn{2}{c}{MNIST}
& \multicolumn{2}{c}{FashionMNIST}
& \multicolumn{2}{c}{SVHN}
& \multicolumn{2}{c}{CIFAR10}
& \multicolumn{2}{c}{ImageNet} \\
\cmidrule(lr){2-3}
\cmidrule(lr){4-5}
\cmidrule(lr){6-7}
\cmidrule(lr){8-9}
\cmidrule(lr){10-11}
\textbf{Model} 
& LeNet-4 & LeNet-5
& Custom-1.6B & Custom-3.3B
& ALL-CNN-A & ALL-CNN-B
& VGG16 & ResNet18
& VGG19 & ResNet50 \\
\midrule
\textbf{Accuracy} 
& 99.04\% & 98.67\%
& 93.45\% & 92.37\%
& 96.17\% & 95.57\%
& 89.43\% & 88.31\%
& 74.22\% & 76.19\% \\
\bottomrule
\end{tabular}
\vspace{-1mm}
\end{table*}

\subsection{Experimental Setup}
\label{experimental}
\noindent\textbf{Target DNN Models and Datasets.} 
Following~\cite{xie2019diffchaser, wang2022bet, lee2020effective, guo2018dlfuzz, kang2020sinvad, dola2024cit4dnn}, we selected five of the most widely used datasets and ten of the most widely adopted DNN models for our experiments.

MNIST~\cite{lecun2010mnist} contains $28\times28$ grayscale images of handwritten digits; it has 60k training inputs and 10k test inputs. We train LeNet-4 and LeNet-5~\cite{lecun1998gradient} on this dataset.

FashionMNIST~\cite{xiao2017fashion} consists of $28 \times 28$ grayscale images of fashion items across 10 categories, with 60k training and 10k test samples. We adopt the FashionMNIST models from Custom~\cite{xiao2017fashion}, using two networks with 1.6 million and 3.3 million parameters, respectively.

SVHN~\cite{netzer2011reading} consists of $32 \times 32$ color digit images cropped from street-view photos, with 73,257 training and 26,032 test samples. We use ALL-CNN-A and ALL-CNN-B models~\cite{springenberg2014striving} for this dataset.

CIFAR10~\cite{krizhevsky2009learning} contains $32\times32$ color images belonging to 10 classes; it has 50k training inputs and 10k test inputs. We used pre-trained VGG16~\cite{simonyan2014very} and ResNet18~\cite{he2016deep}, both fine-tuned for 20 epochs using Adam with a learning rate of $1\times10^{-4}$. 

ImageNet~\cite{deng2009imagenet} contains $224\times224$ color images across 1K classes, with 1.2M training images and 50k validation images. We use pre-trained VGG19~\cite{simonyan2014very} and ResNet50~\cite{he2016deep} as target DNNs and conduct all experiments on the 1K-class ImageNet benchmark.

Table~\ref{tab:models} reports the test accuracies of the target DNN models.
Each dataset is paired with two representative models, all of which achieve strong baseline performance under standard training protocols, ensuring that subsequent testing results are not confounded by undertrained models.

\noindent\textbf{Baselines.} 
We compare \tool\ with representative test generation methods under two oracle settings.
For single-model testing, following prior boundary-oriented studies, we conduct experiments on MNIST, CIFAR10, and ImageNet, and compare \tool\ with SINVAD~\cite{kang2020sinvad} and Mimicry~\cite{weissl2025mimicry}, two latent-space black-box single-model testing approaches.
For multi-model testing, we follow the oracle definition and evaluation protocol of CIT4DNN~\cite{dola2024cit4dnn}, which detects failures via prediction inconsistency across models. Experiments are conducted on MNIST, FashionMNIST, and SVHN, using SINVAD~\cite{kang2020sinvad} and CIT4DNN~\cite{dola2024cit4dnn} as baselines. Mimicry is not included in this setting because it defines failures with respect to a single decision function; extending it to a multi-model oracle would require redefining its failure criterion and deviate from its original design.
Although SINVAD is originally a single-model testing method, it has been adopted under a multi-model oracle in CIT4DNN, where test generation and failure detection are decoupled. We follow this established protocol, and the choice of datasets and baselines is aligned with oracle compatibility and prior evaluations to ensure fair comparison.
In the main comparison, each baseline is evaluated in its original end-to-end configuration, following prior work and preserving the intended workflow of that method. We therefore interpret these results as comparisons between complete testing pipelines rather than fully isolated comparisons of search strategy alone.

\noindent\textbf{Budget.} 
For single-model testing, we uniformly sample correctly classified 100 seeds. For each seed, we generate 10,000 seed-anchor pairs per seed by sampling anchors from labels different from that of the seed according to the anchor-sampling policy described in Section~\ref{sec:3.2}; each anchor defines a latent direction and yields one mutation, total of 10,000 test cases. Baselines use the same seeds and per-seed budget.
For multi-model testing, following the protocol of CIT4DNN, we evaluate all methods under a fixed output budget of 10,000 generated inputs. For \tool, we select seed inputs on which the two models produce identical predictions, then uniformly sample 10,000 seed--anchor pairs and generate one test input from each pair.

\noindent\textbf{Environment.} 
Our experiments were conducted on 64-core PC and NVIDIA RTX 6000 ADA GPU.

\subsection{Metrics}
\label{metrics}
We use the following metrics to measure the effectiveness of the generated test cases. 

\noindent\textbf{\textit{Failure Count}}, as used in previous works~\cite{wang2022bet, xie2019diffchaser, lee2020effective, dola2024cit4dnn}, measures the total number of test cases that trigger prediction discrepancies under the corresponding oracle.

\noindent\textbf{\textit{Seed Coverage}}, as used in previous works~\cite{wang2022bet, lee2020effective}, measures the proportion of input seeds for which at least one failure test case is generated under single-model testing.

\noindent\textbf{\textit{Testing Efficiency}}, following prior work~\cite{lee2020effective,kang2020sinvad,dola2024cit4dnn}, measures the wall-clock time required to generate and evaluate a fixed number of test inputs, reflecting testing-time scalability. Offline training costs are excluded, as all methods rely on pre-trained generative models that are trained once and reused across testing runs.

\noindent\textbf{\textit{Failure Diversity}}, as used in previous works~\cite{wang2022bet, lee2020effective, pei2017deepxplore}, under single-model testing measures the number of distinct alternative predictions predicted by the model for failure-revealing test cases generated from the same input seed. Specifically, for each seed, we count the number of unique predicted outcomes that differ from the model’s original prediction on that seed, and report the average over all seeds.

\noindent\textbf{\textit{Confusion-Pair Diversity}}, under multi-model testing, measures the diversity of disagreement modes revealed by differential testing. For each failure-revealing input where two models produce different predictions, we record the unordered pair of outputs. The metric is defined as the number of distinct confusion pairs observed over all failures, with an upper bound of $\binom{K}{2}$ when the output space contains $K$ discrete categories.

\noindent\textbf{\textit{Semantic Drift}}, following prior works~\cite{caron2021emerging,oquab2023dinov2,darcet2023vision} on visual representation learning, we quantify semantic drift by extracting image representations using a pretrained DINOv2 encoder~\cite{oquab2023dinov2} and computing the cosine distance between the representation of an original seed and that of each corresponding generated test case. DINOv2 is a self-supervised visual model that is widely adopted for measuring semantic variation via representation-space distances. In this work, semantic drift is used as a seed-relative proximity measure.

\section{Results and Analysis}

\subsection{RQ1: Comparison to Single-Model Testing Baselines.}

To assess the behavior of \tool\ under the single-model testing setting, we compare it with representative baselines across five complementary aspects: Failure Count, Seed Coverage, Failure Diversity, Testing Efficiency, and Seed-Relative Semantic Drift. Together, these metrics characterize the fault-exposure capability, efficiency, behavioral diversity, and source-seed proximity of the generated follow-up inputs.
Under the single-model oracle, testing methods start from correctly classified source seeds and generate follow-up inputs, evaluating whether each generated input triggers a prediction discrepancy with respect to the target model. In \tool, each source seed is paired with multiple anchors from alternative classes in the latent space, and each seed-anchor pair defines one controllable mutation direction. This design enables \tool\ to generate diverse follow-up inputs from a single seed through seed-centered, anchor-guided exploration.
To ensure a fair comparison, we randomly select 100 correctly classified source seeds and generate 10,000 test cases per seed for each method. We report mean over 10 independent runs and draw conclusions from consistent performance gaps across datasets and metrics rather than from statistical hypothesis testing alone.

\begin{table*}[t!]
\centering
\small
\setlength{\tabcolsep}{13pt}
\renewcommand{\arraystretch}{0.9}
\caption{Comparison under Single-model Testing Setting}
\vspace{-2mm}
\label{tab:singlemodel}
\begin{tabular}{l l l cc cc cc}
\toprule
\textbf{Dataset} & \textbf{Model} & \textbf{Metric}
& \multicolumn{2}{c}{\textbf{SINVAD}}
& \multicolumn{2}{c}{\textbf{Mimicry}}
& \multicolumn{2}{c}{\textbf{\tool}}\\
\cmidrule(lr){4-5}
\cmidrule(lr){6-7}
\cmidrule(lr){8-9}
& & & Mean & Std & Mean & Std & Mean & Std \\
\midrule
\multirow{6}{*}{MNIST}
& \multirow{3}{*}{LeNet-4}
& $\uparrow$ Failure Count (\#)
& 322 & 28.5 & 2451 & 92.4 & \textbf{2982} & 124.2  \\
&  
& $\uparrow$ Seed Coverage (\%)
& 32 & 2.1 & 99 & 0.4 & \textbf{100} & 0  \\
&  
& $\downarrow$ Testing Efficiency (s)
& 11.8 & 0.8 & 38.3 & 1.4 & \textbf{8.7} & 0.5  \\
\cmidrule(lr){2-9}
& \multirow{3}{*}{LeNet-5}
& $\uparrow$ Failure Count (\#)
& 298 & 24.3 & 2314 & 88.6 & \textbf{2875} & 118.6  \\
&  
& $\uparrow$ Seed Coverage (\%)
& 29 & 1.8 & 99 & 0.3 & \textbf{100} & 0 \\
&  
& $\downarrow$ Testing Efficiency (s)
& 11.4 & 0.8 & 39.4 & 1.2 & \textbf{8.4} & 0.4 \\
\midrule
\multirow{6}{*}{CIFAR10}
& \multirow{3}{*}{VGG16}
& $\uparrow$ Failure Count (\#)
& 684 & 45.2 & 4287 & 124.3 & \textbf{5513} & 168.4 \\
&  
& $\uparrow$ Seed Coverage (\%)
& 43 & 3.4 & 98 & 0.9 & \textbf{100} & 0 \\
&  
& $\downarrow$ Testing Efficiency (s)
& 13.5 & 0.9 & 41.8 & 1.6 & \textbf{9.6} & 0.5 \\
\cmidrule(lr){2-9}
& \multirow{3}{*}{ResNet18}
& $\uparrow$ Failure Count (\#)
& 612 & 41.8 & 4062 & 118.5 & \textbf{4802} & 142.7 \\
&  
& $\uparrow$ Seed Coverage (\%)
& 39 & 2.9 & 98 & 0.7 & \textbf{100} & 0 \\
&  
& $\downarrow$ Testing Efficiency (s)
& 13.1 & 1.1 & 39.2 & 1.5 & \textbf{9.1} & 0.5 \\
\midrule
\multirow{6}{*}{ImageNet}
& \multirow{3}{*}{VGG19}
& $\uparrow$ Failure Count (\#)
& 942 & 62.4 & 5614 & 164.2 & \textbf{6137} & 205.1 \\
&  
& $\uparrow$ Seed Coverage (\%)
& 44 & 4.1 & 98 & 1.3 & \textbf{100} & 0 \\
&  
& $\downarrow$ Testing Efficiency (s)
& 68.4 & 4.5 & 214.3 & 9.4 & \textbf{53.7} & 2.2 \\
\cmidrule(lr){2-9}
& \multirow{3}{*}{ResNet50}
& $\uparrow$ Failure Count (\#)
& 815 & 58.7 & 5429 & 152.6 & \textbf{6284} & 218.4 \\
&  
& $\uparrow$ Seed Coverage (\%)
& 42 & 3.8 & 98 & 1.1 & \textbf{100} & 0 \\
&  
& $\downarrow$ Testing Efficiency (s)
& 72.3 & 5.1 & 198.4 & 8.2 & \textbf{51.3} & 1.9 \\
\bottomrule
\end{tabular}
\vspace{-1mm}
\end{table*}

\begin{figure*}[t!]
  \centering
  \subfloat[MNIST]{
    \includegraphics[width=0.315\linewidth]{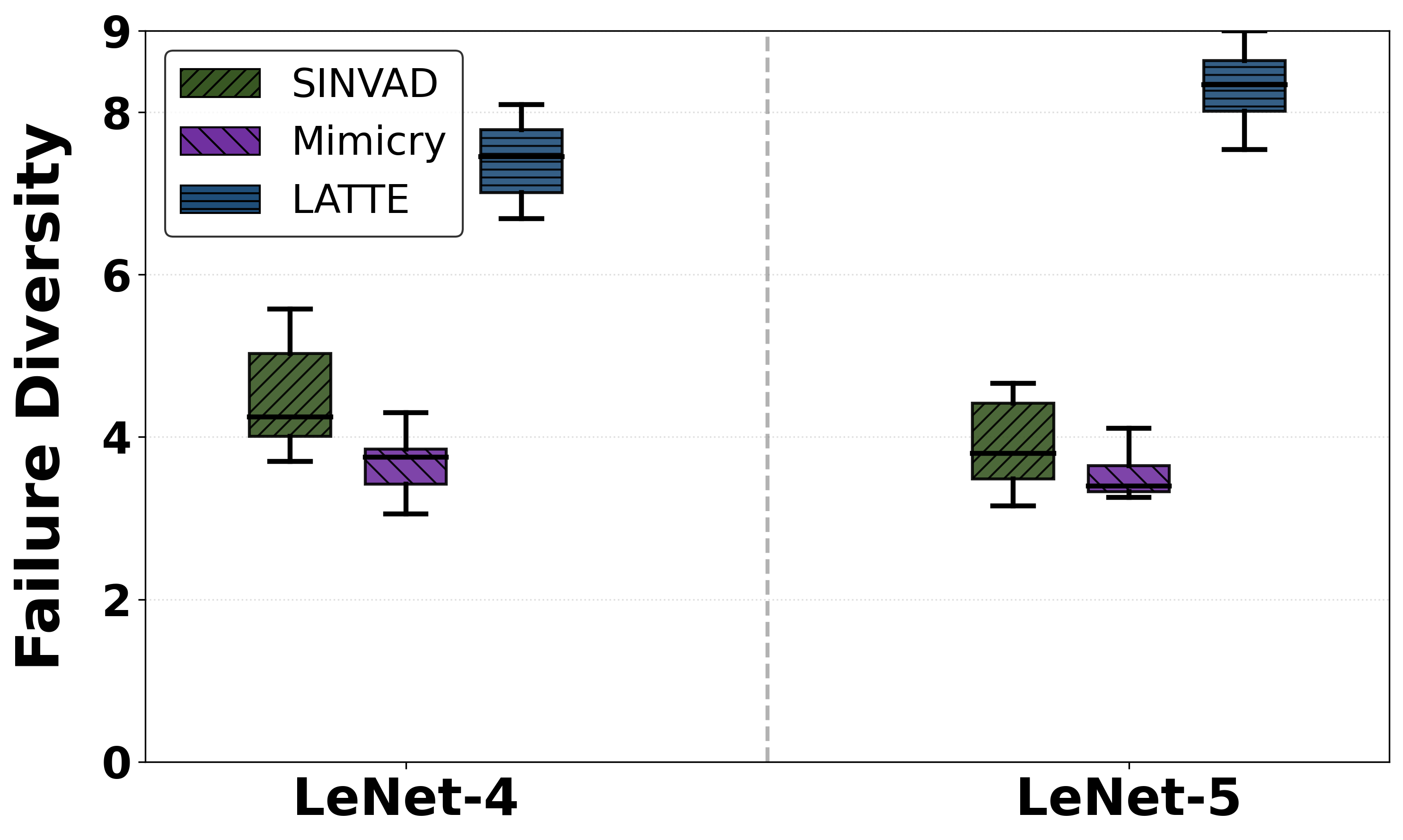}
    \label{fig:boxplot_MNIST_div}
  }
  \subfloat[CIFAR10]{
    \includegraphics[width=0.315\linewidth]{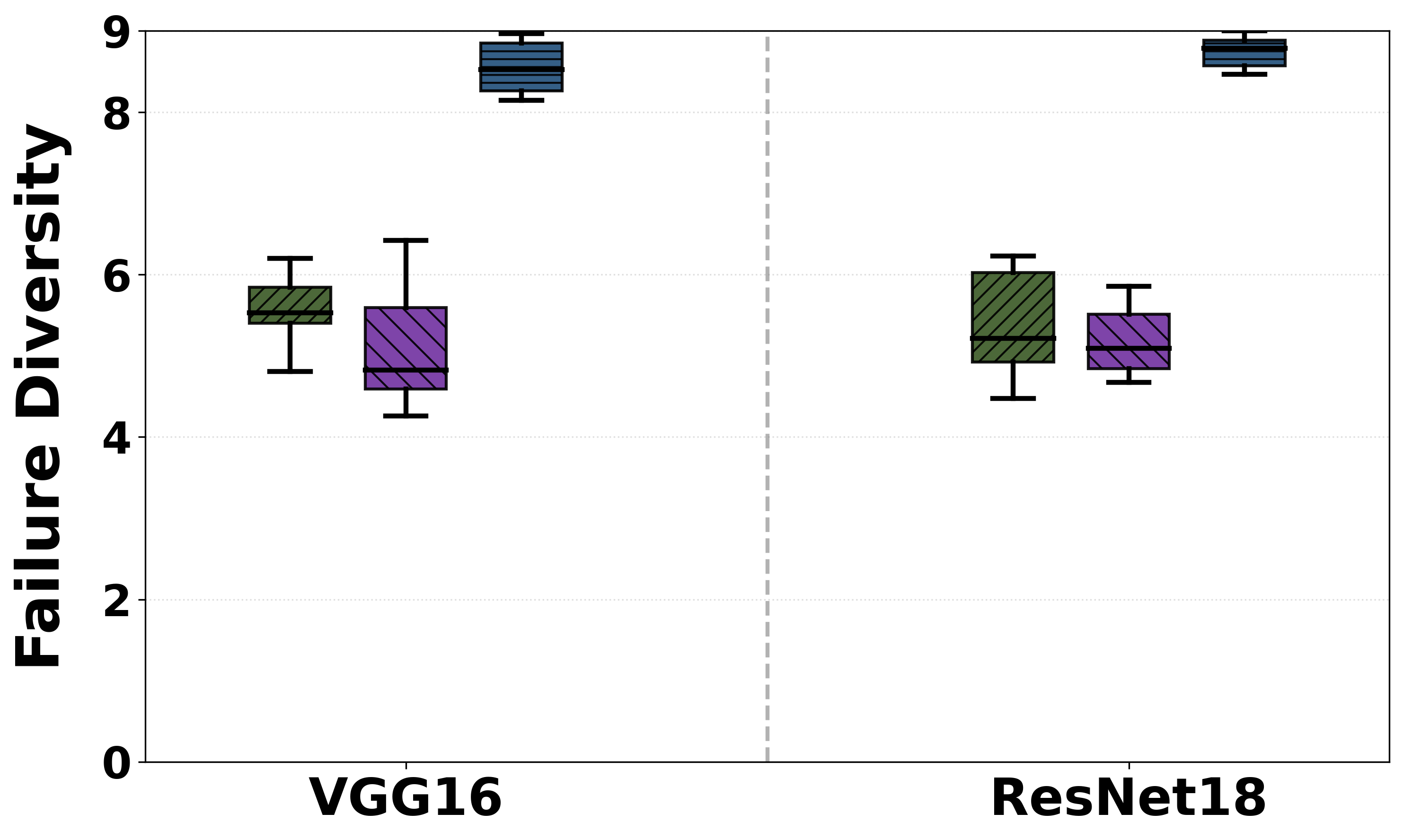}
    \label{fig:boxplot_CIFAR_div}
  }
  \subfloat[ImageNet]{
    \includegraphics[width=0.315\linewidth]{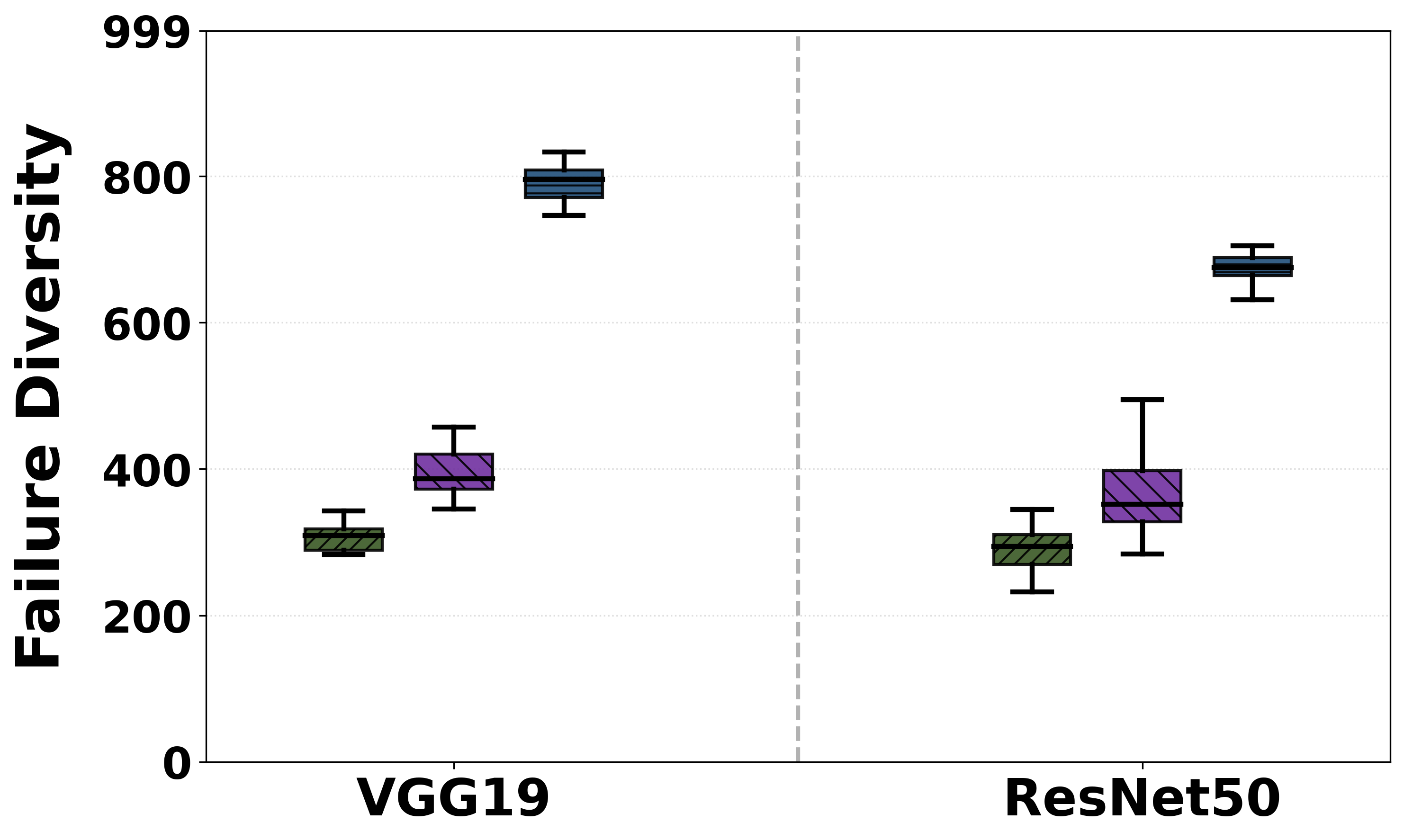}
    \label{fig:boxplot_ImageNet_div}
  }
  \vspace{-1mm}
  \caption{Comparison of Failure Diversity.}
  \vspace{-2mm}
  \label{fig:div_single}
\end{figure*}

\noindent\textbf{\textit{Effectiveness.}}
Table~\ref{tab:singlemodel} summarizes the effectiveness under the single-model testing setting.
Across all datasets and target models, \tool\ consistently generates the largest number of fault-revealing test cases.
On MNIST, \tool\ produces nearly 3,000 failures for both LeNet-4 and LeNet-5, outperforming SINVAD and exceeding Mimicry.
Similar trends are observed on CIFAR10 and ImageNet, where \tool\ discovers over 5,500 failures on VGG16, more than 4,800 on ResNet18, and above 6,000 failures on both VGG19 and ResNet50.

In terms of seed coverage, \tool\ achieves perfect coverage across all datasets and models, ensuring that each input seed yields at least one failure-inducing test case. In contrast, SINVAD covers only 29\%--44\% of seeds, while Mimicry attains high coverage (98\%--99\%).
We note that under the large per-seed budget used here, seed coverage quickly saturates and shows limited variation among strong methods.
We therefore treat seed coverage as a lower-bound indicator that verifies failures are not concentrated on a small subset of seeds, and rely on failure count and diversity for finer-grained comparison.
Overall, under the evaluated single-model settings, \tool\ reveals more failures and does so consistently across seeds, indicating strong fault-exposure capability.

\noindent\textbf{\textit{Efficiency.}}
Table~\ref{tab:singlemodel} reports the testing efficiency under the single-model setting.
Across all datasets and models, \tool\ consistently achieves the lowest wall-clock time under the same budget.
On MNIST and CIFAR10, \tool\ requires only about 8--10 seconds to generate 10,000 test cases, whereas SINVAD typically needs over 11--14 seconds and Mimicry incurs substantially higher overhead, exceeding 38 seconds.
On ImageNet, the efficiency gap becomes more pronounced: \tool\ completes test generation within approximately 51--54 seconds, compared to over 68 seconds for SINVAD and nearly 200 seconds for Mimicry.
Mimicry incurs substantial overhead because it relies on multi-target optimization-driven generation that repeatedly refines candidates with respect to the target model, leading to more iterative computation per test case.
SINVAD is faster than Mimicry but still depends on step-wise, feedback-guided boundary chasing, which requires sequential model queries and limits parallel throughput.
In contrast, \tool\ generates each test case via a single-shot, direction-guided latent mutation with fixed exploration steps, avoiding per-sample iterative refinement and enabling efficient batched decoding and evaluation.
As a result, \tool\ achieves consistently lower generation time, especially on large-scale models such as ImageNet.
These efficiency results reflect online generation-and-evaluation time only and do not include one-time generator training cost.

\noindent\textbf{\textit{Diversity.}}
Figure~\ref{fig:div_single} compares the failure diversity under the single-model oracle.
Note that although Mimicry~\cite{weissl2025mimicry} originally reports ImageNet results on a reduced 10-class subset, here we evaluate all methods on the full ImageNet benchmark to assess their failure diversity under large-scale settings.

Across all datasets and target models, \tool\ consistently achieves the highest failure diversity.
On MNIST and CIFAR10, \tool\ exhibits higher failure diversity than both SINVAD and Mimicry, indicating that the generated failures span a wide range of incorrect predictions rather than collapsing to a few dominant classes.
On ImageNet, where failure diversity is measured over a substantially larger label space, \tool\ uncovers a much broader set of distinct misclassification outcomes, while both SINVAD and Mimicry exhibit more limited coverage.
This trend highlights a fundamental difference in exploration behavior.
SINVAD and Mimicry primarily rely on local or target-specific boundary exploration, which tends to revisit similar misclassification patterns once a boundary is reached.
In contrast, \tool\ explores multiple anchor-guided directions in the latent space, enabling it to traverse diverse decision-uncertain regions around each seed.
As a result, \tool\ not only finds more failures, but also exposes a richer variety of failure modes, and this diversity advantage scales with the number of classes.

\begin{table}[t!]
\centering
\small
\setlength{\tabcolsep}{3.5pt}
\renewcommand{\arraystretch}{0.9}
\caption{Comparison of Seed-Relative Semantic Drift $\downarrow$.}
\vspace{-1mm}
\label{tab:semantic_drift}
\begin{tabular}{l c c c}
\toprule
\textbf{Method}
& \textbf{MNIST}
& \textbf{CIFAR10}
& \textbf{ImageNet} \\
\midrule
\tool\
& \textbf{0.1121} $\pm$ 0.0517
& \textbf{0.1025} $\pm$ 0.0529 
& \textbf{0.0664} $\pm$ 0.0463 \\
SINVAD
& 0.1346 $\pm$ 0.0559
& 0.1239 $\pm$ 0.0546
& 0.0932 $\pm$ 0.0475 \\
Mimicry
& 0.1593 $\pm$ 0.0674
& 0.1418 $\pm$ 0.0597
& 0.1137 $\pm$ 0.0508 \\
\bottomrule
\end{tabular}
\vspace{-2mm}
\end{table}

\noindent\textbf{\textit{Seed-Relative Semantic Drift.}}
To assess how close generated failure-inducing test cases remain to their source seeds, we additionally compare \tool\ with baseline-generated tests in terms of seed-relative semantic drift under the single-model setting. Semantic drift is measured as the representation-level distance between a source seed and its generated follow-up input using a fixed pretrained DINOv2 encoder. Since semantic drift is defined relative to the source seed, this analysis is conducted under the \emph{single-model setting}, where the seed-follow-up relationship is directly central to the testing workflow. For a generated test case $x_{\text{mut}}$ derived from seed $x_{\text{seed}}$, semantic drift is computed as the cosine distance in the representation space:
\[
d(x_{\text{seed}}, x_{\text{mut}}) = 1 -
\frac{\phi(x_{\text{seed}}) \cdot \phi(x_{\text{mut}})}
{\|\phi(x_{\text{seed}})\|_2 \, \|\phi(x_{\text{mut}})\|_2},
\]
where $\phi(\cdot)$ denotes the visual representation extracted by the DINOv2 encoder.

Table~\ref{tab:semantic_drift} reports the mean and standard deviation of semantic drift under matched budgets. Across MNIST, CIFAR10, and ImageNet, \tool\ consistently achieves the lowest semantic drift among the compared methods. This stable ordering across datasets indicates that the follow-up inputs generated by \tool\ remain closer to their source seeds in representation space than those generated by the baselines. Overall, the results suggest that \tool\ maintains stronger seed-relative semantic proximity while generating fault-revealing follow-up inputs.

\noindent\textbf{\textit{Summary.}}
Under the evaluated single-model settings, \tool\ achieves a favorable balance among fault exposure, testing efficiency, behavioral diversity, and seed-relative semantic proximity. Compared with the baseline methods, \tool\ consistently reveals more fault-triggering inputs with near-complete seed coverage, while also requiring less online generation-and-evaluation time. In addition, \tool\ uncovers a broader range of erroneous prediction outcomes and maintains lower semantic drift relative to the source seeds. Taken together, these results indicate that anchor-guided latent exploration is an effective strategy for single-model black-box testing, enabling \tool\ to expose more diverse failures without requiring larger deviation from the original inputs.

\begin{table*}[t!]
\centering
\small
\setlength{\tabcolsep}{8.5pt}
\renewcommand{\arraystretch}{0.9}
\caption{Comparison under Multi-model Testing Setting}
\vspace{-2mm}
\label{tab:multimodel}
\begin{tabular}{l l l cc cc cc}
\toprule
\textbf{Dataset} & \textbf{Models} & \textbf{Metric}
& \multicolumn{2}{c}{\textbf{SINVAD}}
& \multicolumn{2}{c}{\textbf{CIT4DNN}}
& \multicolumn{2}{c}{\textbf{\tool}} \\
\cmidrule(lr){4-5}
\cmidrule(lr){6-7}
\cmidrule(lr){8-9}
& & & Mean & Std & Mean & Std & Mean & Std \\
\midrule
\multirow{2}{*}{MNIST}
& \multirow{2}{*}{LeNet-4 / LeNet-5}
& $\uparrow$ Failure Count (\#)
& 264 & 12.8 & 718 & 41.2 & \textbf{2756} & 95.6 \\
& 
& $\downarrow$ Testing Efficiency (s)
& 11.4 & 0.45 & 9.2 & 0.38 & \textbf{8.7} & 0.52 \\
\midrule
\multirow{2}{*}{FashionMNIST}
& \multirow{2}{*}{Custom-1.6B / Custom-3.3B}
& $\uparrow$ Failure Count (\#)
& 676 & 33.1 & 2696 & 118.4 & \textbf{4726} & 172.8 \\
& 
& $\downarrow$ Testing Efficiency (s)
& 12.7 & 0.52 & \textbf{8.8} & 0.41 & 9.0 & 0.46 \\
\midrule
\multirow{2}{*}{SVHN}
& \multirow{2}{*}{ALL-CNN-A / ALL-CNN-B}
& $\uparrow$ Failure Count (\#)
& 738 & 55.3 & 2463 & 142.6 & \textbf{4991} & 210.4 \\
& 
& $\downarrow$ Testing Efficiency (s)
& 14.3 & 0.68 & 11.2 & 0.55 & \textbf{9.8} & 0.72 \\
\bottomrule
\end{tabular}
\vspace{-2mm}
\end{table*}

\begin{figure*}[t!]
  \centering
  \includegraphics[width=1\linewidth]{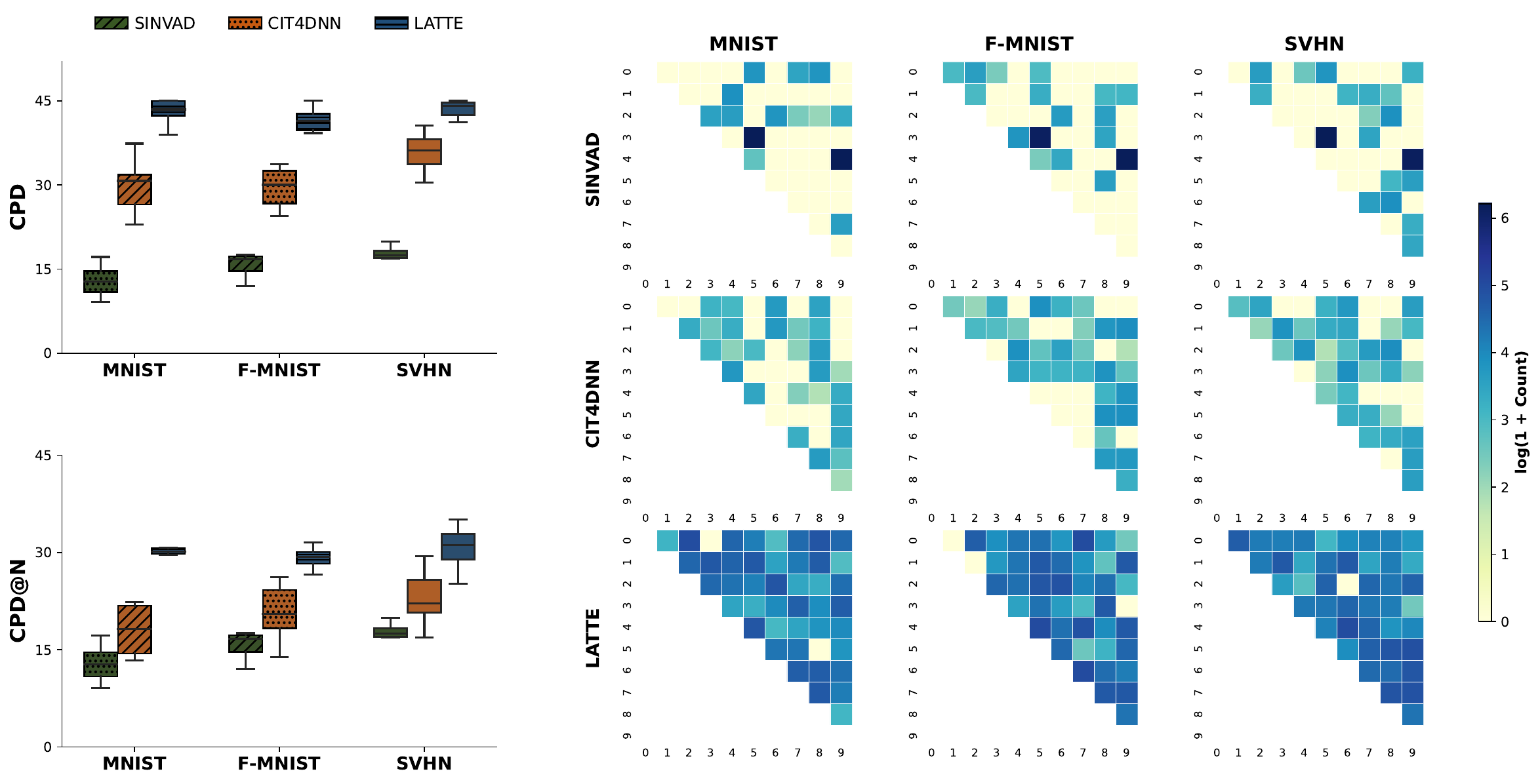}
  \vspace{-4mm}
  \caption{Comparison of Confusion-Pair Diversity}
  \label{fig:multi-div}
  \vspace{-2mm}
\end{figure*}

\subsection{RQ2: Comparison to Multi-Model Testing Baselines.}

To evaluate the behavior of \tool\ under the multi-model testing setting, we compare it with two latent-space-based methods, CIT4DNN~\cite{dola2024cit4dnn} and SINVAD~\cite{kang2020sinvad}, in terms of three key aspects: Failure Count, Testing Efficiency, and Confusion-Pair Diversity. In this setting, a failure is detected whenever two independently trained models produce inconsistent predictions for the same input. Since the multi-model oracle is defined by disagreement between models on a generated input, rather than by inconsistency between a source seed and its follow-up input, we do not separately compare seed-relative semantic drift in this setting. Accordingly, the multi-model comparison focuses on disagreement-triggering capability, efficiency, and disagreement diversity. To ensure a fair comparison, we fix the number of generated test cases to 10,000 for each method and dataset (MNIST, FashionMNIST, and SVHN). Each experiment is repeated 10 times with different random samplings, and we report the average results.

\noindent\textbf{\textit{Effectiveness.}}
Table~\ref{tab:multimodel} summarizes the number of fault-revealing test cases (Failure Count) discovered under the multi-model testing setting. Across all three datasets and model pairs, {\tool} consistently achieves the highest failure count outperforming both CIT4DNN and SINVAD by a clear margin. Specifically, on MNIST, the mean of {\tool} reaches 2,756, which is much higher than CIT4DNN and SINVAD. Similar trends are observed on FashionMNIST and SVHN, where {\tool} uncovers 4,726 and 4,991 failure-inducing cases, respectively. These results demonstrate the capability of {\tool} in exploring the latent space to uncover large number of model inconsistencies.

\noindent\textbf{\textit{Efficiency.}}
Table~\ref{tab:multimodel} compares the generation efficiency of {\tool}, CIT4DNN, and SINVAD. Since all three methods leverage pre-trained generative models to navigate the latent space, they all exhibit high efficiency. Specifically, the time required to generate 10,000 test cases per dataset is consistently within a few seconds on a single GPU. All methods are highly efficient due to batched latent decoding and GPU-parallel evaluation. Minor differences in mean execution time fall within run-to-run variance and do not change the overall efficiency conclusion.


\noindent\textbf{\textit{Diversity.}}
Figure~\ref{fig:multi-div} summarizes the diversity of disagreement patterns revealed under multi-model testing using Confusion-Pair Diversity (CPD). To decouple diversity from the absolute number of detected failures, we additionally report CPD@N, which evaluates confusion-pair diversity under a matched failure budget by randomly sampling with equal probability a fixed number of failure-revealing inputs from each method, set to the minimum observed across methods for each dataset. As shown in the left panels of Fig.~\ref{fig:multi-div}, \tool\ consistently achieves the highest CPD and CPD@N across all datasets, indicating broader coverage of disagreement modes than SINVAD and CIT4DNN even when the number of detected failures is controlled. The right panels visualize how disagreement patterns are distributed across predicted class pairs. For each method and dataset, we count how often different class pairs occur among failure-revealing inputs and visualize these counts as a confusion-pair heatmap. Each heatmap is plotted from one run with 10,000 test cases. Darker colors indicate more frequent class-pair disagreements, while lighter colors indicate rarer patterns. A shared color scale with logarithmic normalization is used to mitigate the influence of a small number of highly frequent pairs and enable comparison across methods. Consequently, concentrated dark regions indicate failures dominated by a small number of disagreement patterns, as observed for SINVAD, whereas \tool\ shows a denser and more evenly spread pattern, indicating broader coverage of disagreement modes.


\noindent\textbf{\textit{Summary.}}
Under the evaluated multi-model settings, \tool\ reveals more disagreement-triggering inputs and covers a broader range of confusion pairs than the compared baselines, while maintaining comparable online testing efficiency. These results indicate that anchor-guided latent exploration is also effective under a differential-testing style oracle.

\subsection{RQ3: Ablation Study}

\subsubsection{Exploration Degree}
\label{sec:RQ4}

To assess the impact of the exploration degree $E$, a key design choice in our approach, we conducted an ablation study. Specifically, for each seed-anchors pair, we use the direction line from seed to anchors in the latent space as the mutation direction and divide this line into 10 equal segments ($E=1$ to $E=10$). We found, when $E$ exceeds 5, the generated test cases become increasingly close to the anchor, diverging from the input seed. Based on these findings, we set the maximum exploration degree to 5 in our main experiments. In this ablation, we focus on two competing factors: fault-revealing capability, measured by Failure Count, and semantic consistency, measured by Semantic Drift. Other metrics, such as diversity, exhibit trends consistent with Failure Count across different exploration degrees, while Seed Coverage shows no clear variation. We therefore concentrate our analysis on these two representative indicators to highlight the trade-off between fault exposure and semantic drift.

\begin{figure*}[t]
    \centering

    \begin{subfigure}{0.325\linewidth}
        \centering
        \includegraphics[width=\linewidth]{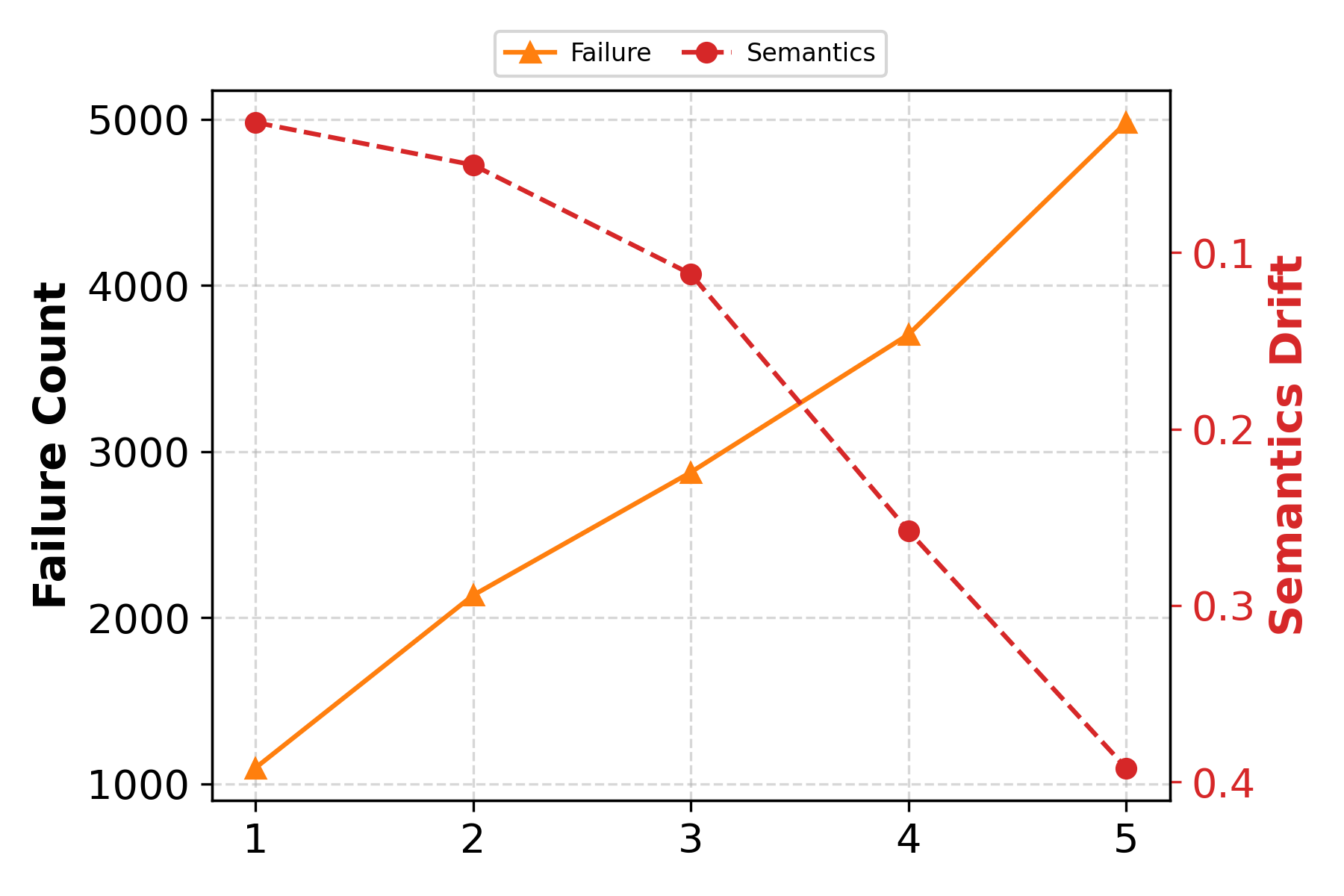}
        \caption{MNIST: LeNet-5}
    \end{subfigure}
    \begin{subfigure}{0.325\linewidth}
        \centering
        \includegraphics[width=\linewidth]{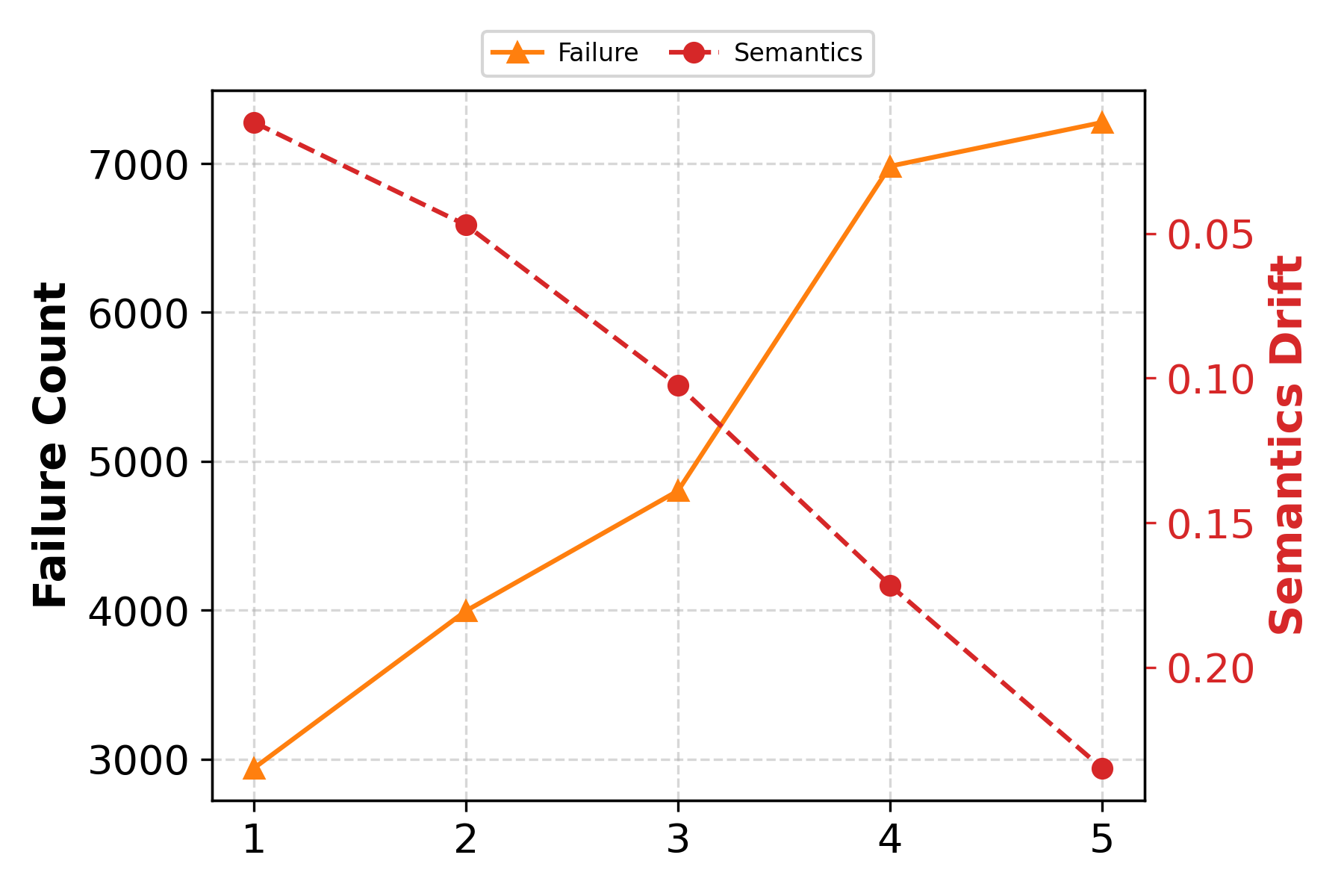}
        \caption{CIFAR10: ResNet18}
    \end{subfigure}
    \begin{subfigure}{0.325\linewidth}
        \centering
        \includegraphics[width=\linewidth]{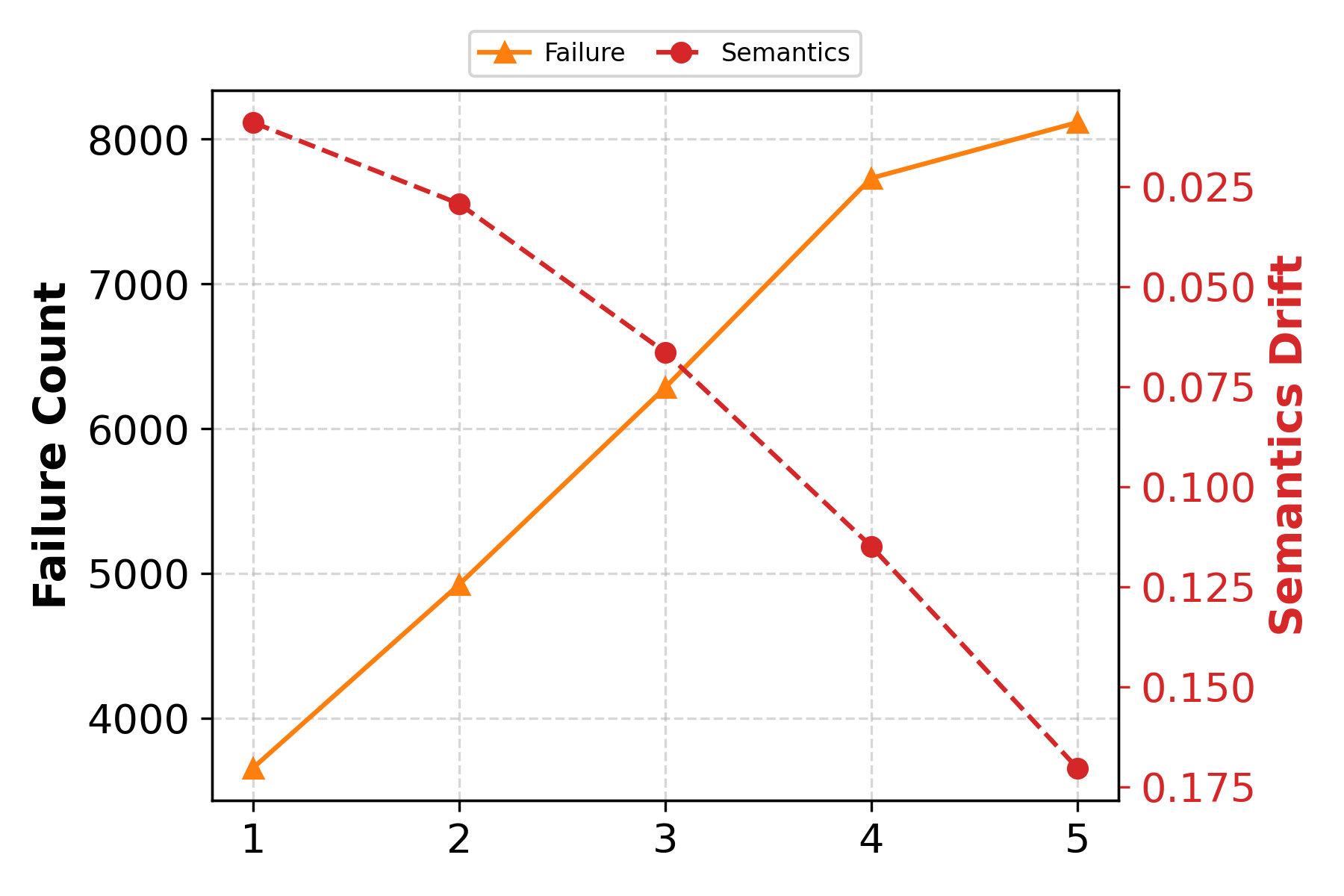}
        \caption{ImageNet: ResNet50}
    \end{subfigure}
    \vspace{-1mm}
    \caption{Trends of Key Results in Single-Model Testing.}
    \label{fig:trend_all}
    \vspace{-4mm}
\end{figure*}

\noindent\textbf{\textit{Single-Model Testing.}}
As shown in Figure~\ref{fig:trend_all}, we examine the impact of exploration degree $E$ (ranging from 1 to 5) on three different models across MNIST, CIFAR10, and ImageNet under the single-model testing setting. We focus on two core metrics: the Failure Count and the Semantic Drift. Following the experimental setup in RQ1, we sample 100 input seeds and generate 10,000 test cases for each. Consistent trends are observed across all datasets and models. As the exploration degree $E$ increases, the Failure Count improves for all models. 
On the other hand, the Semantic Drift exhibits a corresponding upward trend. As $E$ increases from 1 to 5, the semantic distance between the generated test cases and the original seeds grows. Notably, on ImageNet (ResNet-50), the drift remains relatively low (below 0.18 at $E=5$), while on MNIST (LeNet-5), it increases more sharply to approximately 0.29. This highlights an inherent trade-off between test effectiveness and semantic proximity: a more aggressive exploration generates more failures, but excessive drift may eventually diminish the semantic validity of the test cases.In all six models, the growth of Failure Count begins to plateau around $E=3$ or $E=4$, showing diminishing returns in fault-revealing effectiveness. Meanwhile, the Semantic Drift continues to increase, further departing from the seed's original representation. To strike an optimal balance, we select $E=3$ as the default configuration. At $E=3$, \tool achieves high effectiveness, for example, uncovering over 4,300 failures for LeNet-5 and over 6,100 for ResNet-50, while maintaining a reasonable semantic distance (ranging from 0.066 on ImageNet to 0.112 on MNIST). This setting ensures that the generated test cases are both potent in revealing bugs and relatively close to their corresponding original seeds..

\begin{figure*}[t]
    \centering
    \begin{subfigure}{0.325\linewidth}
        \centering
        \includegraphics[width=\linewidth]{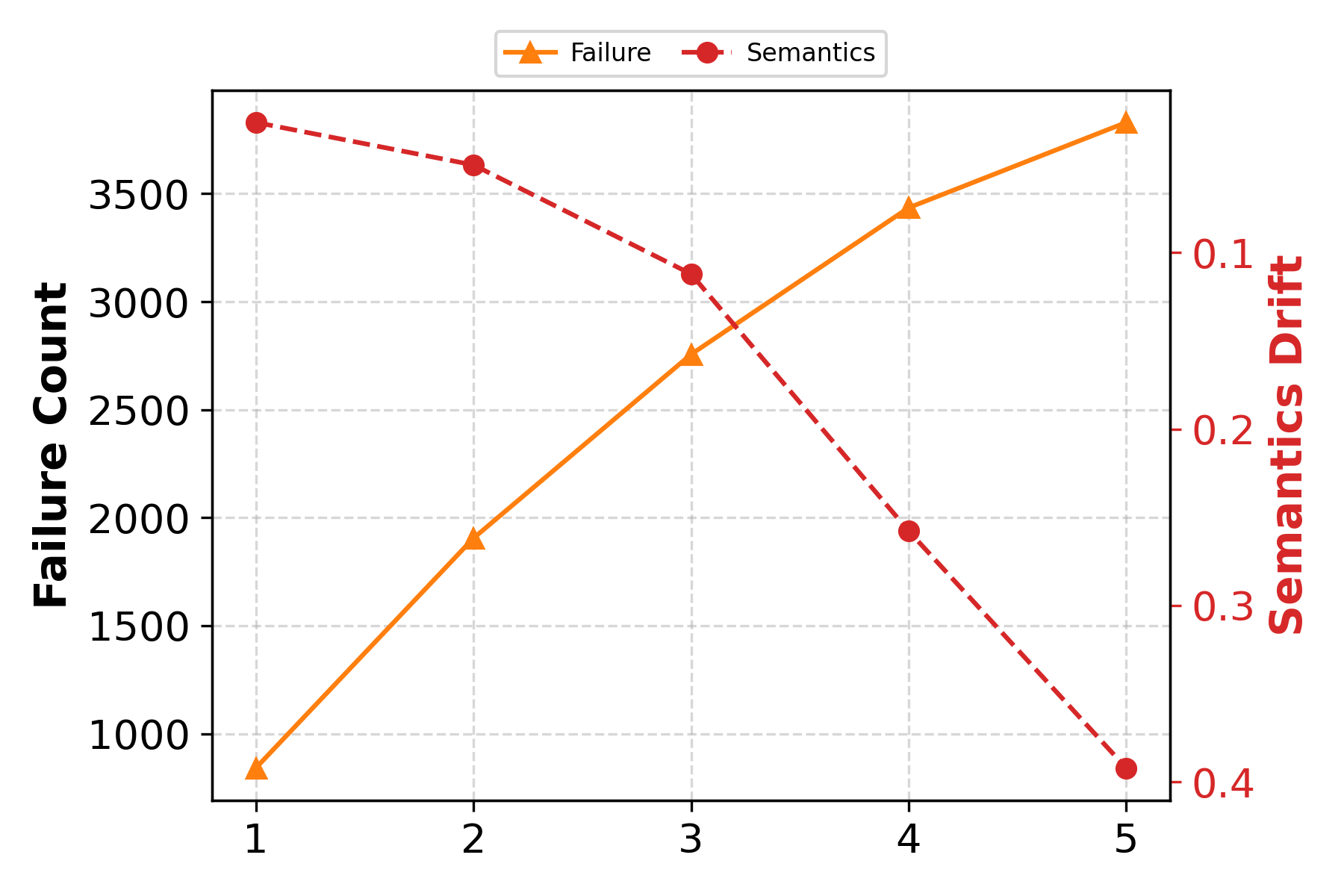}
        \caption{MNIST}
    \end{subfigure}
    \begin{subfigure}{0.325\linewidth}
        \centering
        \includegraphics[width=\linewidth]{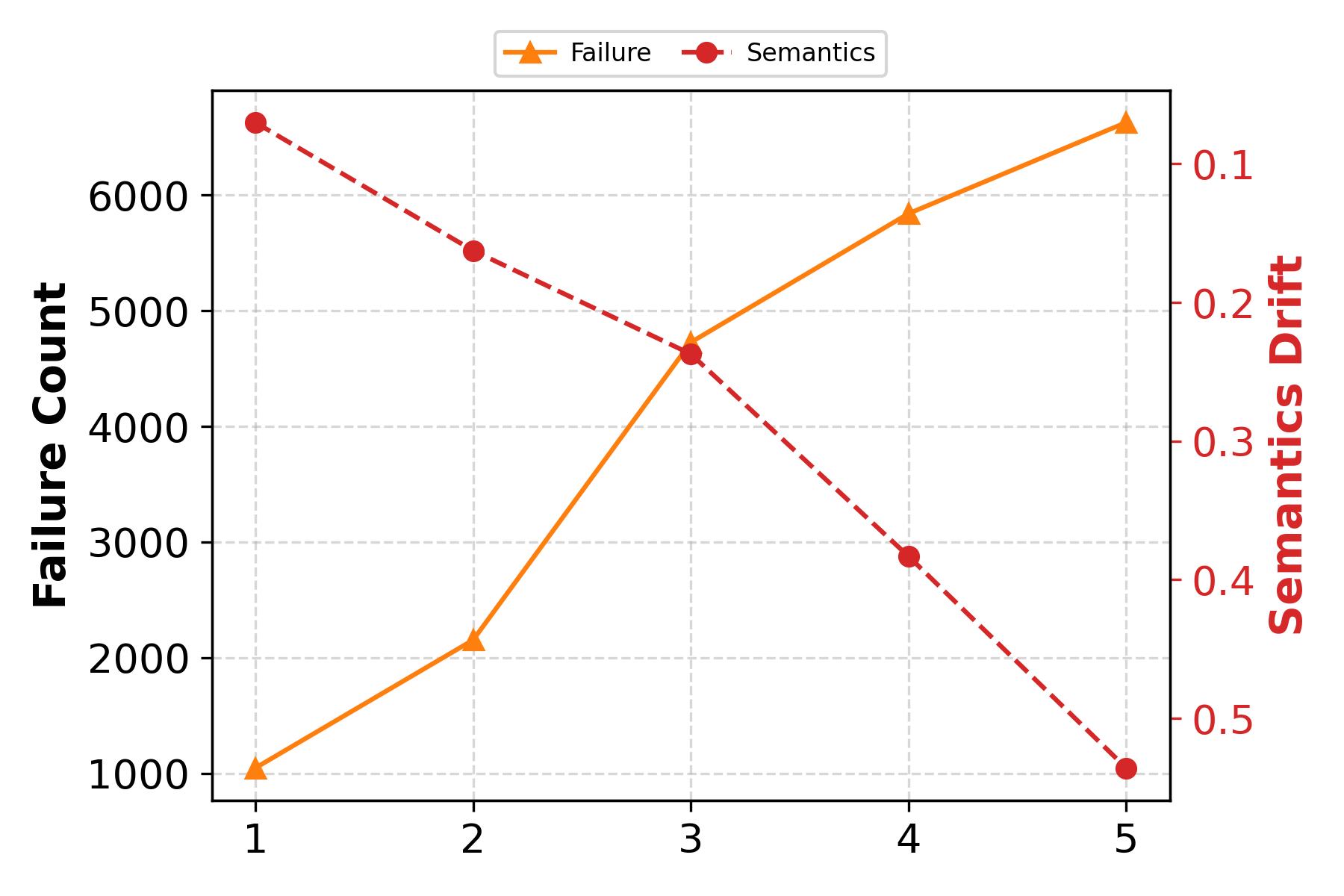}
        \caption{FashionMNIST}
    \end{subfigure}
    \begin{subfigure}{0.325\linewidth}
        \centering
        \includegraphics[width=\linewidth]{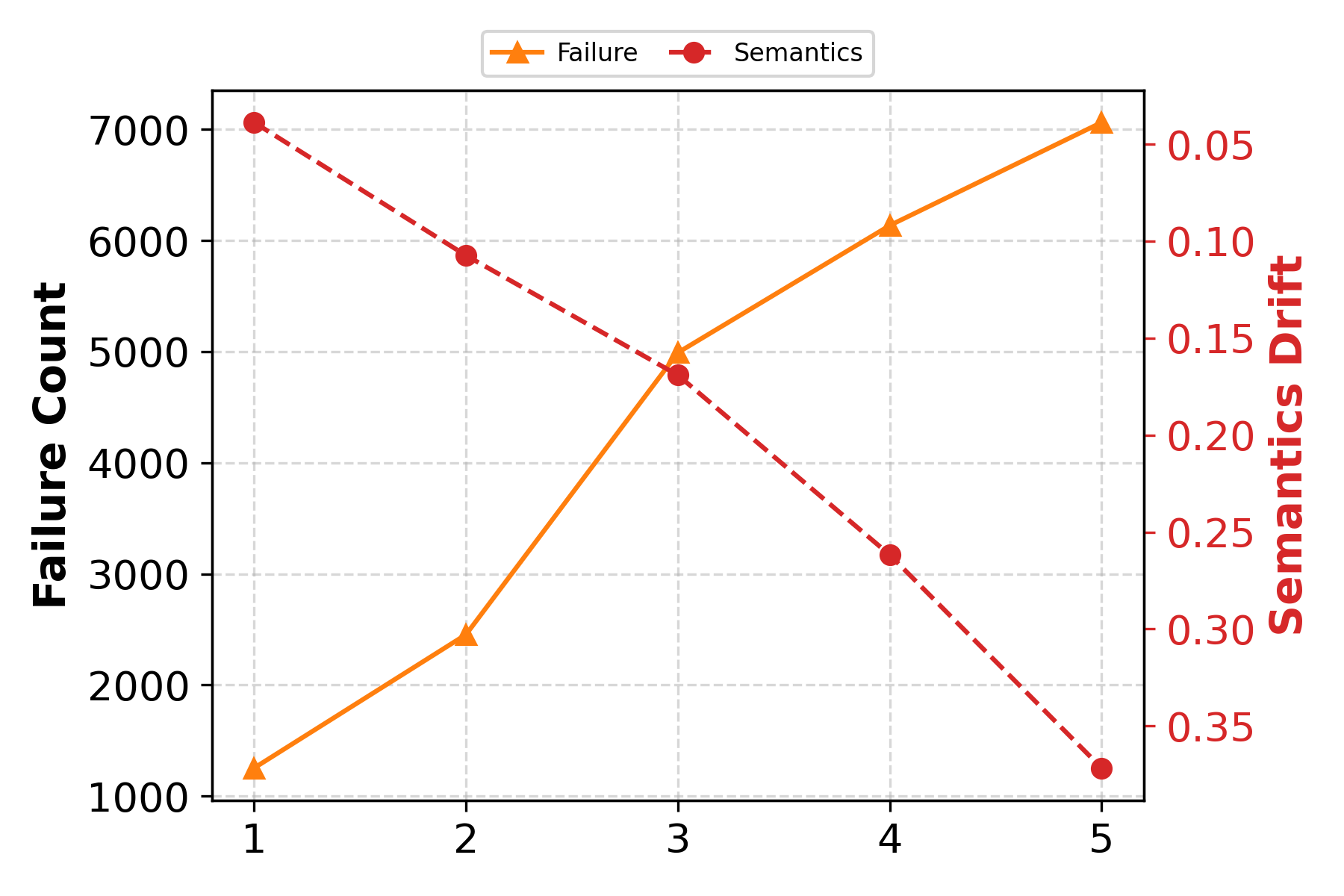}
        \caption{SVHN}
    \end{subfigure}
    \vspace{-1mm}
    \caption{Trends of Key Results in Multi-Model Testing.}
    \label{fig:trend_all_diff}
    \vspace{-4mm}
\end{figure*}

\noindent\textbf{\textit{Multi-Model Testing.}}
As shown in Figure~\ref{fig:trend_all_diff}, all three datasets are under the multi-model testing. Following the experiment setting in RQ2, we generate 10,000 test cases for each dataset.
As the exploration degree $E$ increases from 1 to 5, the normalized Failure (Failure Count) shows a substantial rise for all datasets. This indicates that increasing the degree of latent space exploration enables our method to uncover a greater number of model discrepancies; that is, more fault-revealing test cases are revealed as more regions of the latent space are probed. 
Conversely, Semantic Drift increases with larger exploration degree. For visualization, SD is plotted on a reversed right y-axis. This trade-off highlights that while more in-depth exploration can generate more errors, it also causes test cases to semantically deviate from the original data.
Although $E=2$ yields relatively fewer failures, the generated test cases under the multi-model oracle exhibit limited semantic deviation from the original seeds.
At $E=3$, \tool\ continues to maintain low semantic drift while achieving substantially higher failures count.
We therefore adopt $E=3$ as the default setting, as it best balances fault-revealing capability with semantic consistency.

\subsubsection{Generative Model}
\label{sec:generativemodelablation}

\begin{figure*}[t!]
  \centering
  \includegraphics[width=1\linewidth]{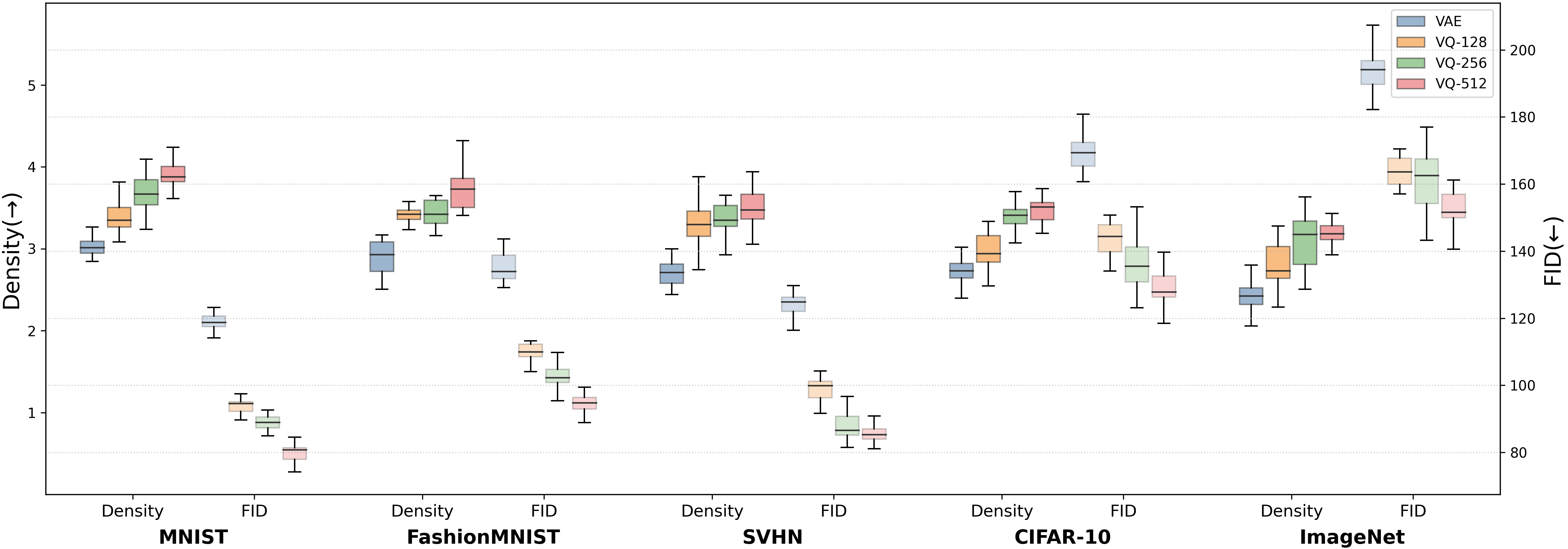}
  \vspace{-4mm}
  \caption{Impact of Different Generative Models on Density and FID.}
  \vspace{-2mm}
  \label{fig:gen_model_ablation}
\end{figure*}

We conduct an ablation study to examine how the choice of generative model affects subsequent testing behavior. We compare a vanilla VAE with three VQ-VAE variants using different codebook sizes (128, 256, and 512) under a fixed exploration degree $E=3$.
All models are trained with matched backbone capacity and identical settings, and results are averaged over 10 independent runs.
In this ablation, we focus on the reconstruction quality and distributional fidelity of different generative models.
Following prior work~\cite{dola2024cit4dnn}, we adopt Fréchet Inception Distance (FID) and Density to assess how well generated samples align with the data distribution induced by the training set.
FID measures global distributional similarity between real and generated samples in the feature space of a pretrained Inception-V3 network:
\(
\mathrm{FID} = \|\mu_r - \mu_g\|_2^2 + \mathrm{Tr}\left(\Sigma_r + \Sigma_g - 2(\Sigma_r \Sigma_g)^{1/2}\right),
\)
where $(\mu_r, \Sigma_r)$ and $(\mu_g, \Sigma_g)$ denote the mean and covariance of real and generated samples, respectively.
As in~\cite{dola2024cit4dnn}, we use FID as a relative proxy for comparing distributional deviation among generative models, rather than as an absolute measure of semantic quality on simple visual domains such as MNIST.
Density, also following~\cite{dola2024cit4dnn}, complements FID by evaluating local reconstruction fidelity, measuring whether generated samples lie in high-density regions of the data manifold using $k$-nearest neighbors ($k{=}5$).
As shown in Fig.~\ref{fig:gen_model_ablation}, all VQ-VAE variants achieve lower FID and higher Density than the vanilla VAE, indicating improved reconstruction quality and a more stable latent space.
Among the tested configurations, VQ-VAE with a codebook size of 512 consistently provides strong performance across datasets, while larger codebooks introduce higher variance and less stable reconstruction.
Therefore, we adopt VQ-VAE (512) as a balanced generative model.

\section{Discussion}

\begin{figure*}[t!]
  \centering
  \includegraphics[width=1\linewidth]{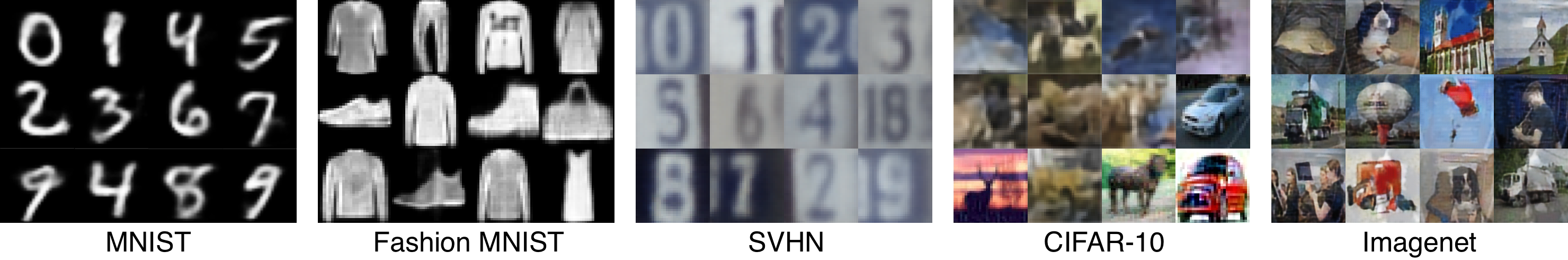}
  \vspace{-4mm}
  \caption{Example of \tool\ generated test cases.}
  \vspace{-2mm}
  \label{fig:example}
\end{figure*}

\noindent\textbf{Case Study.}
Figure~\ref{fig:example} shows representative follow-up inputs generated by \tool\ across five datasets (MNIST, FashionMNIST, SVHN, CIFAR10, and ImageNet). These examples illustrate that \tool\ can generate visually plausible test inputs in the latent space and decode them back into the input domain under different image settings. This qualitative impression is also consistent with the seed-relative semantic drift observed for \tool, particularly on the more visually complex datasets. For example, on CIFAR10 and ImageNet, \tool\ yields low mean semantic-drift values in representation space, suggesting that many generated follow-up inputs remain close to their corresponding source seeds despite being fault-revealing.

\begin{figure}[t!]
  \centering
  \subfloat[Entropy distribution]{
    \includegraphics[width=0.48\linewidth]{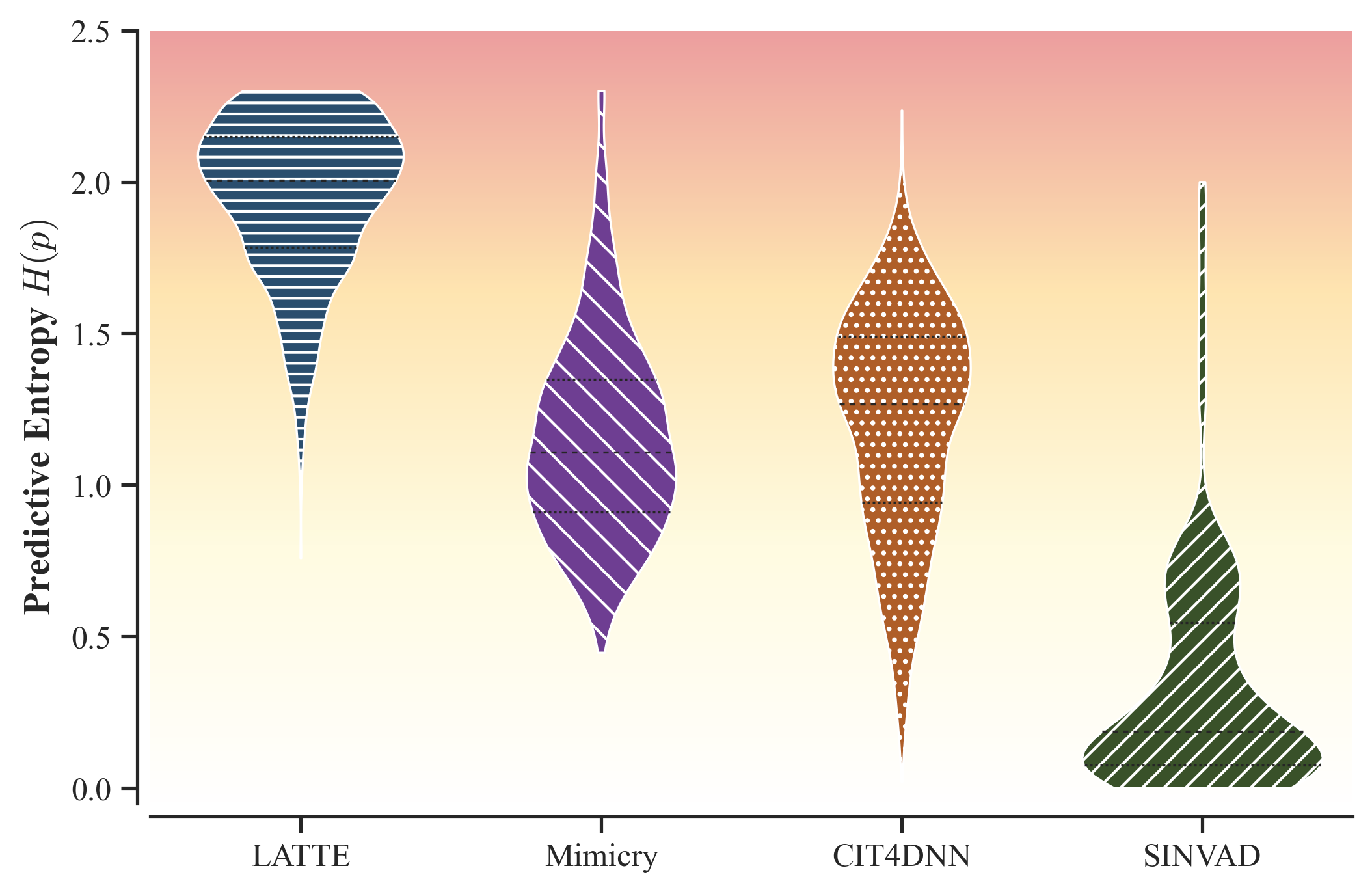}
    \label{fig1}
  }
  \subfloat[Entropy evolution]{
    \includegraphics[width=0.48\linewidth]{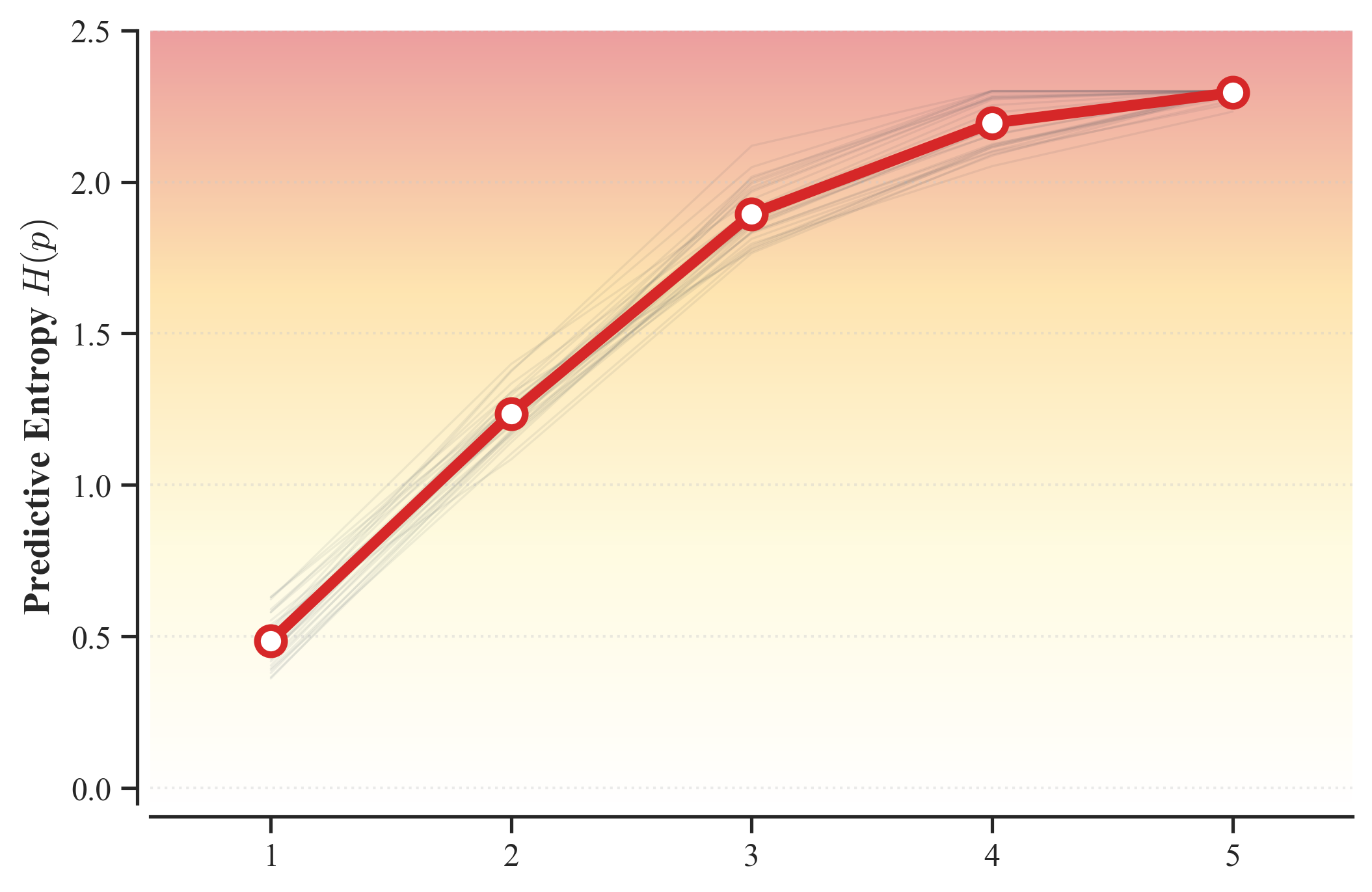}
    \label{fig2}
  }
  \vspace{-1mm}
  \caption{Uncertainty Analysis of Generation Behavior.}
  \vspace{-3mm}
  \label{fig:discussion}
\end{figure}

\noindent \textbf{Uncertainty Proximity Analysis.}
Although \tool\ does not explicitly optimize predictive uncertainty, we perform a post-hoc analysis on MNIST using softmax predictive entropy to characterize the uncertainty of the generated test cases.
Fig.~\ref{fig1} shows that \tool\ produces an entropy distribution consistently shifted toward higher values than the baselines, suggesting that its generated inputs more frequently reside in low-confidence prediction regions.
Fig.~\ref{fig2} further illustrates the entropy evolution along seed$\rightarrow$anchor exploration paths: predictive entropy typically increases at intermediate steps and gradually saturates as the exploration approaches the anchor.
This oracle-independent analysis does not imply a causal relationship between predictive entropy and failure detection.
Rather, it provides empirical evidence that, on MNIST, anchor-driven latent exploration tends to traverse regions associated with elevated predictive uncertainty, supporting the design motivation of \tool.

\noindent \textbf{Comparison with adversarial attacks.}
Our work, in line with prior neural network testing methods~\cite{pei2017deepxplore,guo2018dlfuzz,lee2020effective,wang2022bet,dola2024cit4dnn}, differs fundamentally from adversarial attacks~\cite{chakraborty2021survey,huang2017adversarial,villegas2024evaluating}. Adversarial attacks typically seek minimal, often imperceptible input-space perturbations under an explicit perturbation budget and are primarily evaluated by attack success rate. In contrast, \tool\ aims to generate oracle-triggering test inputs that expose diverse failure modes under a fixed testing budget. Methodologically, adversarial attacks explore local input-space neighborhoods, whereas \tool\ explores latent neighborhoods defined by semantically meaningful anchors, motivated by evidence that learned latent representations capture meaningful and diverse directions of variation~\cite{kingma2013auto,higgins2017beta,voynov2020unsupervised}. Accordingly, adversarial attacks are not used as baselines in our evaluation, and we instead evaluate \tool\ using testing-oriented metrics such as fault exposure, behavioral diversity, and semantic drift.

\noindent \textbf{Future work.}
Our current evaluation focuses on image classification. Extending \tool\ to other tasks will require task-specific generators and oracle definitions, which we leave to future work.

\section{Threats to Validity}

\noindent \textbf{Internal Validity.}
We carefully reviewed the implementation to ensure result stability and followed standard training protocols for all models, including pretrained backbones and original configurations.
Our single-model oracle treats prediction inconsistency as a testing signal rather than a definitive semantic error.
Although such inconsistencies may arise near ambiguous predictions, we mitigate invalid mutations by constraining test generation within the latent space and by evaluating seed-relative semantic proximity using representation-based metrics under the single-model setting.

\noindent \textbf{External Validity.}
Our evaluation focuses on image classification tasks and a set of representative DNN architectures.
While these settings are consistent with prior DNN testing studies, the generalizability of \tool\ to other tasks or model families remains an open question.

\section{Related Work}

DNN testing approaches generate follow-up test inputs by exploring either the input space, a feature space, or a learned latent space.
DeepXplore~\cite{pei2017deepxplore}, DeepTest~\cite{tian2018deeptest}, DLFuzz~\cite{guo2018dlfuzz}, ADAPT~\cite{lee2020effective}, and BET~\cite{wang2022bet} work on the input space and generate test inputs by applying pixel-level transformations on seed inputs. The diversity of the generated tests is limited by the diversity of the seed inputs used by these methods, and they operate in the input space, where only significant perturbations affect classification, often deviating from the original data semantics.

Feature-space test generation methods create in-distribution test cases but rely on a model that captures the characteristics of the input data distribution. When such a model is available, approaches like DeepHyperion~\cite{zohdinasab2021deephyperion} and DeepJanus~\cite{riccio2020model} have proven effective in exploring the feature space and uncovering faults. However, constructing such models demands considerable domain expertise and is especially challenging in high-dimensional scenarios like DNNs.
In contrast, latent-space generators such as SINVAD~\cite{kang2020sinvad}, CIT4DNN~\cite{dola2024cit4dnn}, and Mimicry~\cite{weissl2025mimicry} generate tests by navigating the latent space of a learned generator, which typically yields more in-distribution cases than input-space mutations. However, these methods are primarily driven by latent coverage or optimization objectives, rather than by seed-centric exploration of neighborhoods.
SINVAD performs unconstrained updates in a continuous VAE latent space, which can repeatedly revisit near-duplicate regions or drift into semantically misaligned areas due to the lack of explicit directional control.
CIT4DNN discretizes a continuous latent into factors and allocates budget to satisfy $t$-way combinatorial coverage; while systematic, this spreads queries across many factor combinations and reduces per-seed focus on uncertainty regions.
Mimicry formulates test generation as iterative multi-objective optimization, where exploration is guided by optimizing explicit trade-offs and requires multiple refinement steps, biasing the search toward specific optima.

In contrast, \tool\ frames latent-space test generation as seed-centric, anchor-guided directional exploration: each seed is paired with anchors from alternative classes, and each pair yields one controllable mutated latent point. This design emphasizes multi-direction exploration around the source seed rather than unconstrained search, factor-coverage objectives, or iterative optimization.

\section{Conclusion}

In this paper, we presented \tool, a seed-centric, anchor-guided latent-space framework for black-box testing of image classifiers. \tool\ generates follow-up inputs by mutating each original seed along directions defined by anchors from alternative classes and decoding the mutated latent points back into the input space. This design enables controllable multi-direction exploration around each seed under both single-model and multi-model testing oracles.
Our evaluation across five datasets and ten DNN models shows that, under the evaluated settings, \tool\ improves fault exposure and behavioral diversity compared with representative latent-space baselines. At the same time, semantic-drift analysis under the single-model setting indicates that many \tool-generated tests remain close to their original seeds in representation space. Taken together, these results suggest that anchor-guided latent exploration is a promising strategy for black-box testing of image classifiers when fault exposure, behavioral diversity, and seed-relative semantic proximity must be balanced under fixed testing budgets.

\bibliographystyle{IEEEtran}
\bibliography{main1}
\end{document}